\pdfoutput=1
\documentclass[11pt]{article}

\usepackage[final]{acl}

\usepackage{times}
\usepackage{latexsym}

\usepackage[T1]{fontenc}
\usepackage[utf8]{inputenc}

\usepackage{microtype}

\usepackage{inconsolata}

\usepackage{graphicx}

\usepackage{booktabs}
\usepackage{dsfont}
\usepackage{microtype}
\usepackage{graphicx}
\usepackage{array}
\usepackage{booktabs}
\usepackage{multirow}
\usepackage{multicol}
\usepackage{color}
\usepackage{threeparttable}
\usepackage{amsmath}
\usepackage[utf8]{inputenc}
\usepackage{longtable}
\usepackage{amssymb}
\usepackage{makecell}
\usepackage{xspace}
\usepackage{xcolor}
\usepackage{colortbl}

\usepackage{kotex}
\usepackage{tabularx}
\usepackage{graphicx}
\usepackage{todonotes}
\usepackage{listings}

\definecolor{Gray}{gray}{0.94}

\definecolor{lightgreen}{RGB}{194, 245, 186}
\definecolor{darkgreen}{RGB}{80, 182, 39}
\definecolor{lightblue}{RGB}{163, 223, 255}

\definecolor{green7}{RGB}{160,220,160}
\definecolor{green6}{RGB}{180,230,180}
\definecolor{green5}{RGB}{200,240,200}
\definecolor{green4}{RGB}{220,250,220}
\definecolor{green3}{RGB}{235,255,235}
\definecolor{green2}{RGB}{245,255,245}
\definecolor{green1}{RGB}{252,255,250}

\newcommand{\minisection}[1]{%
\vspace{0.04in}
    \noindent \textbf{#1}.\xspace%
}

\newcommand{\ours}{\texttt{LLM-AT}\xspace}
\newcommand{\oursacc}{\textit{accuracy estimator}\xspace}

\makeatletter
\def\@fnsymbol#1{\ensuremath{\ifcase#1\or \dagger\or
   1\or \mathparagraph\or \|\or **\or \dagger\dagger
   \or \ddagger\ddagger \else\@ctrerr\fi}}
\makeatother

\newcommand*\samethanks[1][\value{footnote}]{\footnotemark[#1]}

\title{Automatic Transmission for LLM Tiers: \\Optimizing Cost and Accuracy in Large Language Models}

\author{Injae Na\textsuperscript{\thanks{\ \ Major in Bio Artificial Intelligence}}, Keonwoong Noh\textsuperscript{\samethanks[1]}, \and Woohwan Jung \\
  Department of Applied Artificial Intelligence, Hanyang University \\
  \texttt{\{suhoij47, rohgw011, whjung\}@hanyang.ac.kr} }

\begin{document}
\maketitle
\begin{abstract}
LLM providers typically offer multiple LLM tiers, varying in performance and price. 
As NLP tasks become more complex and modularized, selecting the suitable LLM tier for each subtask is a key challenge to balance cost and performance. 
To address the problem, we introduce the \textbf{LLM A}utomatic \textbf{T}ransmission (\ours) framework that automatically selects LLM tiers without training. %\footnote{\url{https://github.com/hyudsl/LLM-AT}}
\ours consists of Starter, Generator, and Judge. 
The starter selects the initial LLM tier expected to solve the question, the generator produces a response using the LLM of the selected tier, and the judge evaluates its validity.
If the response is invalid, \ours iteratively upgrades to a higher-tier model, generates a new response, and re-evaluates until a valid response is obtained.
Additionally, we propose \oursacc, allowing the selection of a suitable initial tier without training.
Given an input, \oursacc estimates the expected accuracy of each LLM tier by computing the valid response rate for top-k similar queries from past inference records. 
Experiments demonstrate that \ours achieves superior performance while reducing costs, making it a practical solution for real-world applications. 
Our code is available at \url{https://github.com/hyudsl/LLM-AT}.

\end{abstract}

\section{Introduction}

\begin{figure}[tb]
  \includegraphics[width=\columnwidth]{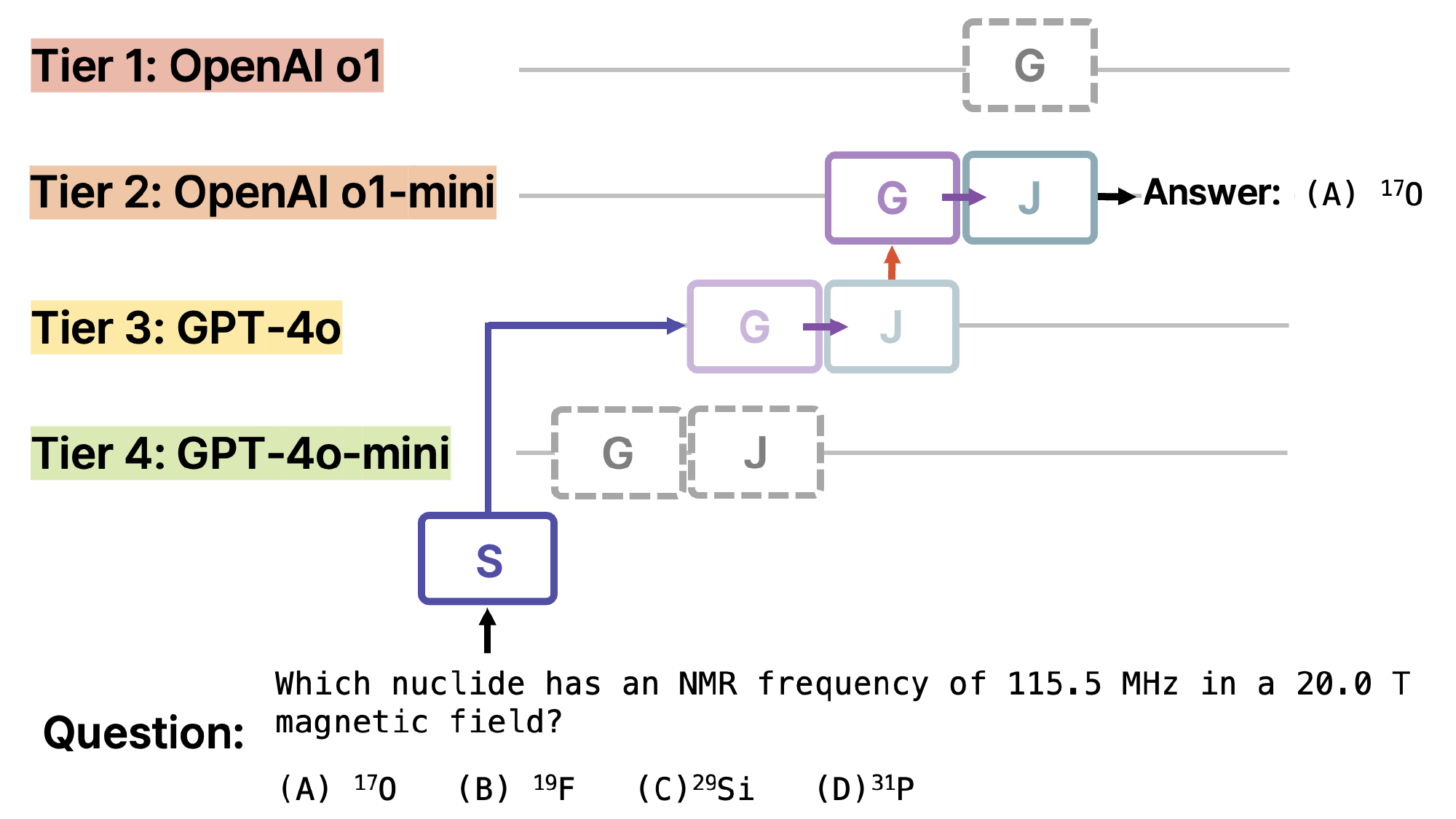}
  \caption{
    An example of the \ours process with an input question. `S', `G', and `J' indicate the Starter, Generator, and Judge, respectively.
  }  
  \label{fig:example_of_llm_at}

\end{figure}

Large Language Models (LLMs) have become powerful tools for a wide range of natural language tasks.
%~\cite{achiam2023gpt} 
Due to the high computational demands for serving large-scale LLMs, LLM providers such as OpenAI typically offer multiple tiers of proprietary models (e.g,  4o-mini, 4o, o1-mini, and o1). %\cite{o1-mini, o1}. 
Higher-tier models exhibit superior performance but come at a higher cost, while lower-tier models are cheaper but fall short in accuracy.
 
As NLP tasks are more complex~\cite{lan-etal-2024-llm, fu-etal-2024-msi, zong-etal-2024-triad}, tasks are solved by dividing them into multiple subtasks, each requiring different levels of reasoning or generation capabilities.  
For instance, the Tree of Thoughts approach \cite{yao2024tree} often solves a single complex task with more than tens of LLM calls.
This introduces a new challenge: selecting the suitable LLM tier for each substask to optimize the trade-off between accuracy and cost.
Although using the top-tier model for all subtasks guarantees high-quality results, it can be prohibitively expensive when a complex task requires a lot of LLM calls.
Conversely, relying solely on lower-tier models reduces costs but risks performance degradation. 

Recent studies~\cite{ding2024hybrid, chen2024frugalgpt, ong2024routellm} have explored training-based approaches to select the suitable LLM, but have several limitations.
First, these methods require extensive data labeling to assess whether each LLM generates accurate responses to a given question, making the process time-consuming and costly.
Second, when new LLMs are released, retraining is necessary to integrate new LLMs into the selection process.
Third, these methods become unreliable when applied to test domains that differ from the training data distribution,  limiting their generalizability in real-world applications.

To address this problem, we propose the \textbf{LLM A}utomatic \textbf{T}ransmission (\ours), a novel framework that dynamically selects the suitable LLM tier without training.
\ours consists of three main components: Generator, Judge, and Starter.
The generator produces a response for a given question using the initial LLM tier selected by the starter.
The judge then evaluates the validity of the response. If the response is invalid, \ours upgrades to the higher tier LLMs and repeats the process until a valid response is obtained.
Figure~\ref{fig:example_of_llm_at} shows an example of the \ours process. Using GPT-4o  selected as the initial tier by the starter, the generator answers a given question.
However, the judge determines that the response is invalid, \ours upgrades to the higher-tier model, o1-mini, which then produces the valid output.

Starting with a low LLM tier for a complex question can cause unnecessary tier escalations, leading to increased costs and execution time.
Therefore, we introduce \oursacc that estimates the probability of each LLM tier successfully answering a given question by leveraging real-time accumulated inference records, referred to as History, without requiring correctness-labeled data.
This enables the starter to select the cheapest LLM tier expected to generate a correct response while reducing additional LLM calls.
Our main contributions are summarized as follows:
\begin{itemize}
    \item{We propose the \ours framework, which automatically selects the LLM tier for different-level tasks.}
    \item{We present an \oursacc that predicts the probability of each LLM tier answering correctly without correctness-labeled data.}
    \item{We conduct extensive experiments across various question difficulty levels, ranging from simple to graduate-level tasks. We demonstrate that \ours optimizes the trade-offs between accuracy and monetary costs, as well as accuracy and execution time.}
\end{itemize}

\section{Related Works}

\minisection{LLM Routing}
% 학습방법을 위주로 글을 전개해보자.
Several studies have explored learned LLM routing methods to select the most suitable LLM for a given question.
\citet{chen2024frugalgpt} employs a cascading mechanism, where LLMs are sequentially invoked using a stop function that evaluates response quality.
However, since this method always starts with the cheapest model from a fixed sequential permutation, complex queries necessitate multiple model invocations, leading to increased cost and latency.
\citet{ding2024hybrid, ong2024routellm} formulate routing as a binary classification problem, directing simple queries to smaller models and complex ones to larger models.
However, these approaches fail to account for the diverse performance levels of proprietary LLMs, which are rapidly emerging, leading to suboptimal routing decisions and reduced flexibility in real-world applications.
On the other hand, \citet{lu-etal-2024-routing, chen2024routerdc, maurya2024selectllm} explore routing as an alternative to ensembling multiple LLMs.
Thus, these methods focus on select the best-performing LLM on a given question, disregarding inference cost.

\minisection{Iterative Refinement in LLMs}
Several studies have explored iterative refinement techniques to enhance the accuracy and reliability of LLM-generated responses. 
These approaches enable models to evaluate their own outputs, incorporate feedback, and iteratively improve their responses without additional training.
\citet{madaan2023selfrefine} introduce a method where LLMs generate self-feedback on their initial responses and refine them through multiple iterations, improving factual consistency and coherence.
\citet{gou2024critic} propose an iterative verify-then-correct framework that enhances model reliability by using external tool-assisted critiques.
\citet{yao2023react} presents a prompting strategy that incorporates reasoning traces, allowing LLMs to dynamically track, update, and adjust their responses.
Existing works show that LLMs can engage in self-evaluation and iterative refinement, leveraging both internal reasoning and self-feedback to enhance response quality over multiple iterations.

\begin{figure*}[t]

  \includegraphics[width=1\linewidth]{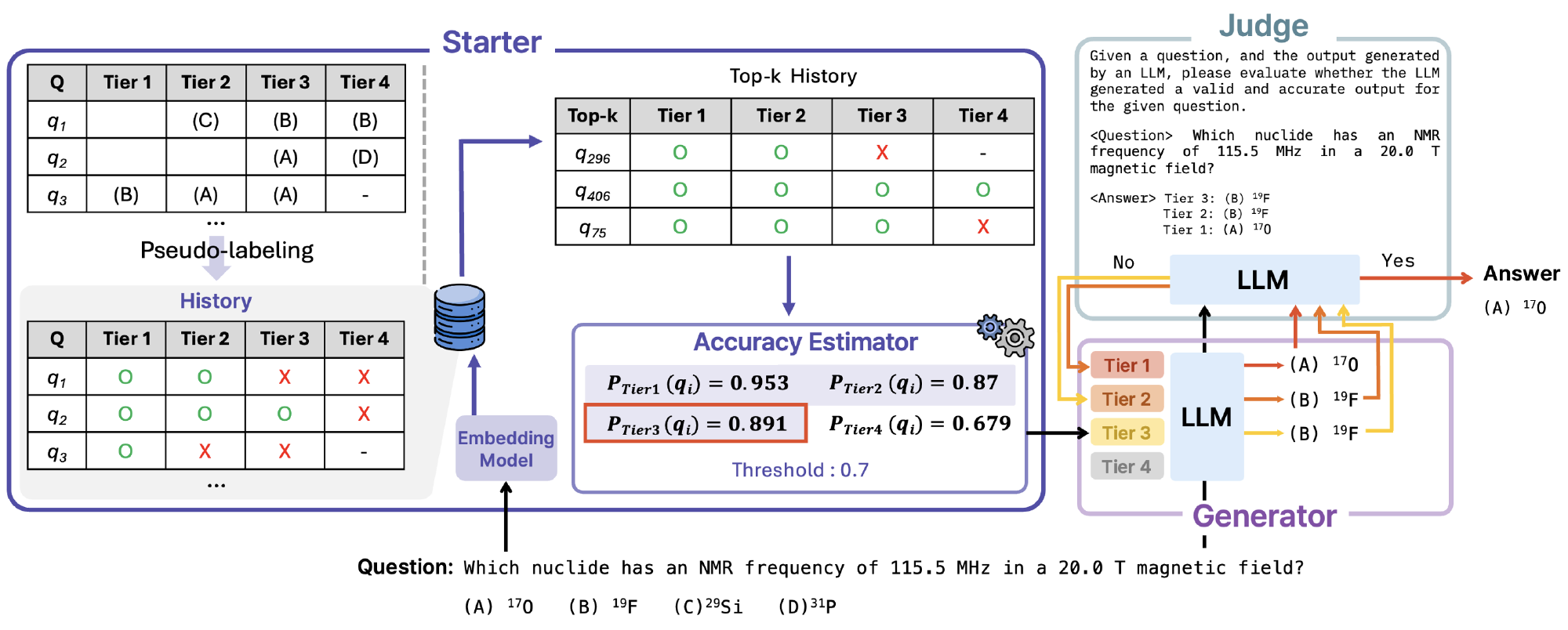}
  \caption {An overview of \ours Framework.}
  \label{fig:Framework}
\end{figure*}

\section{\ours Framework}
%\todomemo{여기서부터 다시 다듬는 중/ input: question, 답변: answer로 되도록 통일}

Most LLM providers offer various tiers of LLMs with different price points and performance levels.
In general, higher-tier models provide better performance but come with higher usage costs.
Therefore, this study aims to optimize the trade-off between performance and cost given the LLM tiers.
Specifically, our goal is to achieve performance comparable to the top-tier model while minimizing overall costs, including monetary expenses and execution time, when handling inputs of varying difficulty levels. 
Additionally, we assume an environment where past interactions can be stored and utilized.
It is a realistic scenario widely adopted in actual LLM-based chatbots such as ChatGPT~\cite{Chatgpt}, Claude~\cite{Claude} and DeepSeek~\cite{DeepSeek} etc.

% While \ours can be applied to any tier-based LLM systems, such as ChatGPT and Claude, we will illustrate our method with the tier system of ChatGPT as an example: 4o-mini \cite{4o-mini}, 4o \cite{4o}, o1-mini \cite{o1-mini}, and o1 \cite{o1}.

\subsection{Overview}
\label{overview}
Figure~\ref{fig:Framework} shows an overview of the proposed \ours framework.
The starter, which determines the initial tier, consists of a history that stores the questions processed by each tier model along with the corresponding correct and incorrect pseudo-labels for the questions, and an accuracy estimator that calculates the estimated accuracy of each tier based on the history.
Once the starter selects the initial tier, the corresponding generator produces an answer, and the judge evaluates its validity. 
If the generated answer is invalid, as shown in tier 3 of Figure~\ref{fig:Framework}, the tier is upgraded by one level, and the process repeats with the new generator and judge.
If the generator produces a valid answer, as shown in tier 2, that response is returned as the final answer. 
The process continues until a valid answer is generated or the system reaches the top-tier model.
The input to the generator varies depending on the task, but here, we define it as a question.
While \ours can be applied to any tier-based LLM systems, such as ChatGPT and Claude, we will illustrate our method with the tier system of ChatGPT as an example: o1, o1-mini, GPT-4o, and GPT-4o-mini.

\subsection{Generator}
\label{Generator}

Rather than focusing on enhancing the performance of a single model, we investigate methods to leverage multiple tiers of models to achieve improved responses effectively.
Therefore, to minimize the model dependency, we employ CoT~\cite{NEURIPS2022_9d560961} and PoT~\cite{chen2022program} prompting, which have been shown to be effective across a range of LLMs, rather than relying on model-specific prompting methods.
For detailed prompts, please refer to the appendix~\ref{appendix:prompts}. 

Our system, as explained in Section~\ref{overview}, performs a type of iterative refinement process, where inference is repeated within the given tier system until a valid answer is generated.
Although providing the output of lower-tier to higher-tier can enhance performance ~\cite{gou2024critic, madaan2023selfrefine}, we avoid this approach due to the potential risk of inaccurate responses from lower tier that impede accurate generation of higher tier and increasing input token processing costs.

\subsection{Judge}
Inspired by~\citet{gou2024critic}, we use the judge module to assess the answer of the generator.
The judge is composed of an LLM of the same tier as the generator and receives the question and the answer of the generator, evaluates its validity and outputs either \texttt{‘yes’} or \texttt{‘no’}.
Generating CoT reasoning along with the answer can potentially assist the judge’s evaluation.
However, the judge can leverage the reasoning process generated by the generator for answer evaluation, and especially since the cost of generating output tokens is higher than processing input tokens, we prompt the judge to provide concise answers without CoT reasoning to minimize the output token cost. 
Additionally, using external tools could assist in the validity evaluation, but we do not use them to ensure that the method works well in general cases.
For example, integrating external tools such as RAG or search APIs could improve the accuracy of the judge.

\subsection{Starter}
Starting from the lowest tier and sequentially moving upward can cause unnecessary API cost and execution time when high-tier models are needed to answer a question. 
Furthermore, low-tier LLMs still have limitations in self-evaluation~\cite{huang2023large}, leading to performance degradation.
If we use a high-tier judge model to improve performance, it can significantly increase evaluation costs.
To address this problem, we propose the starter, which starts from an appropriate tier for a given question rather than always beginning from the lowest tier. 
This method reduces cost, time, and API calls while mitigating overestimation risks at lower tiers.

To select the appropriate initial tier for a given question, the starter estimates the accuracy of each tier for the question and then selects the lowest-cost model among those exceeding a specific threshold as the initial tier model.
If no tiers exceed the threshold, it defaults to starting from the lowest tier.
To estimate the accuracy of each tier, it is possible to directly use the publicly available benchmark performance of each tier. 
However, this performance reflects the average results across the entire benchmark and does not account for the individual difficulty of each question.
Since the accuracy of each tier can vary depending on the difficulty of the question, we propose \oursacc, a method that estimates the accuracy which considers the difficulty and topic of the question.

In this section, we first introduce a pseudo-labeling method to annotate the correctness labels of previous answers by the LLMs.
Then, we present the accuracy estimation methods to calculate the accuracy of each tier.

% S: Preprocessing
% Pseudo-labeling Preprocessing
% Answer Evaluation Preprocessing
% Postprocessing
\subsubsection{Pseudo-labeling of Correctness}
To estimate the accuracy of each tier, we rely on previous inference records and the correctness labels.
Ideally, human annotated labels would be used, but they are usually unavailable.
Instead, we pseudo-label the correctness according to the following assumptions based on the validation by the judges:
\begin{itemize}
  \setlength{\itemsep}{-1pt}     % space between items
  \setlength{\topsep}{0pt}      % space before/after the list
    \item If the judge evaluates the response as valid, the answer is labeled as correct.
    \item If a given LLM tier is correct, we assume any higher tier model also generates correct answers. 
    For example, in $q_1$ of Figure~\ref{fig:Framework}, tier 2 generates a valid answer, so tier 1 is not used; however, the pseudo label still receives the correct label.
    \item If a lower-tier generates the same answer as the valid answer, that tier is labeled as correct; otherwise, it is labeled incorrect. 
    Therefore, tier 3 and 4, which generate answers different from tier 2 in $q_1$, are assigned incorrect labels.
    \item For lower tiers skipped by the starter, we leave the labels blank, as there are no results and evidence to evaluate, just like tier 4 in $q_3$.
\end{itemize}
We store the correctness labels with the corresponding questions in a database, referred to as History, for the accuracy estimation.
%The correctness label for each tier is stored with the question in a database to be used in accuracy estimation.
%We refer to the database as History.

% S: Accuracy estimation
\subsubsection{Accuracy Estimation}
%In order to determine the most suitable tier for the current question, we calculate the accuracy of each tier using their pseudo-labels. 
Training a dedicated accuracy estimation model would require significant labeled data, as well as additional training and usage costs. 
To avoid these overheads, we propose a simpler, counting-based approach that relies on correctness labels in the history DB. 
Specifically, we count how many times each tier responded correctly or incorrectly for similar queries and use these counts to estimate the tier accuracy.

When calculating the accuracy of each tier for a question $q$, the \oursacc uses the pseudo-labels of the top-$k$ most similar questions in the history based on the cosine similarity of their embedding vectors.
By focusing on similar questions, the estimator more accurately reflects how each tier might perform on $q$.

The estimated accuracy $P_j(q)$ of tier $j$ is computed by dividing the number of correct labels by the number of all labels  as
\begin{equation}
\label{eq1}
P_j(q)= \frac{n^T_{j} +\alpha^T}{n^T_{j}+n^F_{j}+\alpha^T+\alpha^F}  
\end{equation}
where  $n^T_{j}$ and $n^F_{j}$ are the number of correct and incorrect responses among the $k$ answers, respectively.
$\alpha^T$ and $\alpha^F$ are hyperparameters for the prior distributions \cite{jung2019crowdsourced}.
Note that we use a benchmark score to utilize the background knowledge on the performance of the LLMs.

\begin{equation}
\alpha^T=\lambda \cdot Acc^{Bench},\;\alpha^F = \lambda\cdot (1-Acc^{Bench})
\end{equation}
where $\lambda$ is a hyperparameter to adjust the importance of the background knowledge and $Acc^{Bench}$ is the benchmark score.
If benchmark performance is not available, the alpha is set to 1.
%To compute more stable accuracy values, a smoothing factor $\alpha$~\cite{feng2024diverseagententropy} is added.

% n에 대한 설명 

\minisection{Similarity-aware accuracy estimation}
To give more importance to the questions that are highly similar to $q$, we weight the contribution of each question by its similarity $\mathrm{sim}(q, q\prime)$
 instead of simply counting the frequency.
Thus, the number of correct answers $ n^T_{j}$, and that of incorrect answers $n^F_{j}$ for tier $j$ among the top-k questions, are calculated as follows.
\begin{equation}
\label{eq2}
n^T_{j}\!=\! \sum\limits_{q\prime\in\\top(q)}  \mathrm{sim}(q, q\prime) \!\cdot\! \mathds{1}(l_{j,q\prime}\ \text{is correct})
\end{equation}
\begin{equation}
\label{eq3}
n^F_{j}\!=\! \sum\limits_{q\prime\in\\top(q)} \mathrm{sim}(q, q\prime) \!\cdot\! \mathds{1}(l_{j,q\prime}\ \text{is incorrect})
\end{equation}
where $top(q)$ is the set of top-k similar questions of $q$ and $l_{j, q{\prime}}$ is the correctness label of tier $j$ for $q{\prime}$.

\section{Experimental Settings}

\subsection{Datasets}
We use two datasets with varying difficulty of the question to evaluate whether \ours effectively selects the LLM tier while optimizing trade-offs between performance, cost, and time.

\minisection{MATH} The MATH~\cite{hendrycksmath2021} comprises challenging mathematical competition problems organized into five difficulty levels.
Level 1 contains the easiest questions, whereas level 5 contains the most difficult ones.
To reflect the real-world scenario where easier questions are more prevalent, we sample 400 test examples from MATH, starting with 100 samples from Level 1 and decreasing by 10 per level as difficulty increases.

\minisection{MCQA} We construct the Multiple Choice Question Answering (MCQA) dataset by combining the following five datasets with different difficulty levels: OpenBookQA~\cite{OpenBook}, ARC-Easy, ARC-Challenge~\cite{allenai:arc}, MMLU~\cite{mmlu}, GPQA~\cite{rein2024gpqa}.
The easiest level consists of 500 elementary-level science questions sampled from the OpenBookQA test set. 
The second and third levels comprise 400 and 300 questions from ARC-Easy and ARC-Challenge, respectively, covering science topics for students in grades 3 to 9. 
The fourth level includes 200 questions from MMLU, spanning elementary to advanced professional levels. 
The hardest level contains 100 Ph.D.-authored questions from GPQA.

%------
\subsection{LLM Tier System} We use the OpenAI tier system consisting of the following four models\footnote{The tier system is as of 2025-1-17}: o1, o1-mini, GPT-4o, and GPT-4o-mini.
Table~\ref{tab:price_of_model_tier} shows the price of LLMs.

\begin{table}[tb]
\centering
\small
\begin{tabular}{lcrr}
\toprule
%\textbf{Model} & \textbf{Tier} & \multicolumn{2}{c}{\textbf{Price (\$/1k tokens)}} \\
\multirow{2.5}{*}{\textbf{Model}} & \multirow{2.5}{*}{\textbf{Tier}} 
& \multicolumn{2}{c}{\textbf{Price (\$/1M tokens)}} \\
\cmidrule(lr){3-4}
 & & \textbf{Input} & \textbf{Output} \\ 
\midrule
o1          & 1 & 15.00 & 60.00 \\
o1-mini     & 2 & 3.00 & 12.00 \\
GPT-4o      & 3 & 2.50 & 10.00 \\
GPT-4o-mini & 4 & 0.15 & 0.60 \\
\bottomrule
\end{tabular}
\caption{Pricing for LLMs in OpenAI tier system.}
\label{tab:price_of_model_tier}
\end{table}

As an exception, we provide the abstention option for the lowest tier, GPT-4o-mini, allowing complex problems to be forwarded to higher tiers thus reducing the risk of hallucination and improving accuracy~\cite{Abstain}.
If the GPT-4o-mini abstains, the system upgrades directly to the next tier without going through the judge, saving both the output token cost of the generator and the cost of the judge.
In addition, we set the judge to GPT-4o, which is one tier higher than GPT-4o-mini, to avoid overestimation and ensure evaluation accuracy.
% If the GPT-4o-mini abstains, the system upgrades directly to the next tier without going through the judge, saving both the generator’s output token cost and the judge’s cost

\subsection{Compared Methods}

We compare our proposed method with two inference strategies: single inference (Single) and iterative inference (Iteration).

\minisection{Single} The single inference follows the standard LLM inference approach, where the model generates an answer in one pass. 

\minisection{Iteration} Following~\citet{madaan2023selfrefine}, the iterative inference allows an LLM to refine its answer through up to three cycles of generation and validation, stopping early if a valid answer is found.
If no valid answer is produced by the last cycle, the output of the last cycle is used as the final answer.

\subsection{Evaluation Metric}
We evaluate \ours and the baselines in terms of API cost, execution time, and accuracy.
First, API cost is measured in dollars, based on the token price of OpenAI as shown in Table~\ref{tab:price_of_model_tier}.
Second, execution time is measured in minutes as the elapsed time for running inference over the entire dataset.
Finally, the performance of the model is evaluated using the accuracy metrics originally defined in the datasets that comprise MATH and MCQA.

\subsection{Implementation Details of \ours}
% Alibaba-NLP/gte-Qwen2-1.5B-instruct
We apply a few-shot PoT prompt to the MATH dataset and a zero-shot CoT prompt to the MCQA dataset. 
In the case of MATH using PoT, we utilize a Python interpreter to obtain the results of the generated code.
Both datasets use the top-5 questions most similar to the input question to select the initial tier, with a threshold of 0.7 and the value of $\lambda$ set to 5.
For top-k similar question retrieval, we used the embedding model \textit{Alibaba-NLP/gte-Qwen2-1.5B-instruct} from Hugging Face\footnote{https://huggingface.co/Alibaba-NLP/gte-Qwen2-1.5B-instruct/}, with cosine similarity as the similarity metric.
% \footnote{https://huggingface.co/Alibaba-NLP/gte-Qwen2-1.5B-instruct}

Furthermore, we measure the accuracy of each tier in MMLU Pro \cite{wang2024mmlupro}, a new dataset not included in our test set, and use these values as the $Acc^{Bench}$ values.
This is done to ensure generalization testing and fair comparison, and using the benchmark performance suited for each dataset leads to significantly better results. 
The experimental results for this are detailed in Appendix~\ref{specailized factor}, and the more detailed experimental setting can be found in Appendix~\ref{detail setting}.

\section{Experimental Results}

\begin{table*}[t]
\centering
\resizebox{0.9\linewidth}{!}{%
\begin{tabular}{lcccc|cccc}
\toprule 
& \multicolumn{4}{c|}{\textbf{MATH}} 
& \multicolumn{4}{c}{\textbf{MCQA}} \\
\cmidrule(lr){2-5} \cmidrule(lr){6-9}
& \multicolumn{3}{c}{\textbf{Judge}} & \textbf{Generator}
& \multicolumn{3}{c}{\textbf{Judge}} & \textbf{Generator} \\
\cmidrule(lr){2-4} \cmidrule(lr){5-5} \cmidrule(lr){6-8} \cmidrule(lr){9-9}
\textbf{Model}& \textbf{F1 score} & \textbf{Recall} & \textbf{Precision} & \textbf{Accuracy}
& \textbf{F1 score} & \textbf{Recall} & \textbf{Precision} & \textbf{Accuracy} \\
\midrule
GPT-4o-mini & 0.763 & 0.943 & 0.641 & 0.531 & 0.918 & 0.948 & 0.890 & 0.863 \\
GPT-4o-mini* & 0.828 & 0.877 & 0.784 & 0.531 & 0.939 & 0.974 & 0.906 & 0.863 \\
GPT-4o      & 0.799 & 0.858 & 0.748 & 0.610 & 0.943 & 0.976 & 0.912 & 0.894 \\
o1-mini     & 0.876 & 0.954 & 0.811 & 0.749 & 0.942 & 0.970 & 0.915 & 0.897 \\
\bottomrule
\end{tabular}%
}
\caption{Performance of the judge and generator on MATH and MCQA. Results marked with * use the special judge (GPT-4o). TP: correct and valid; FP: incorrect but valid; FN: correct but invalid; TN: incorrect and invalid.}
\label{tab:judge performance}
\end{table*}

\subsection{Main Results}
\label{sec:main}
\begin{figure}[t]
    \centering
    \hspace{-1cm}
    \setlength{\abovecaptionskip}{5pt}
    \setlength{\belowcaptionskip}{1pt}
    \resizebox{\columnwidth}{!}{\includegraphics{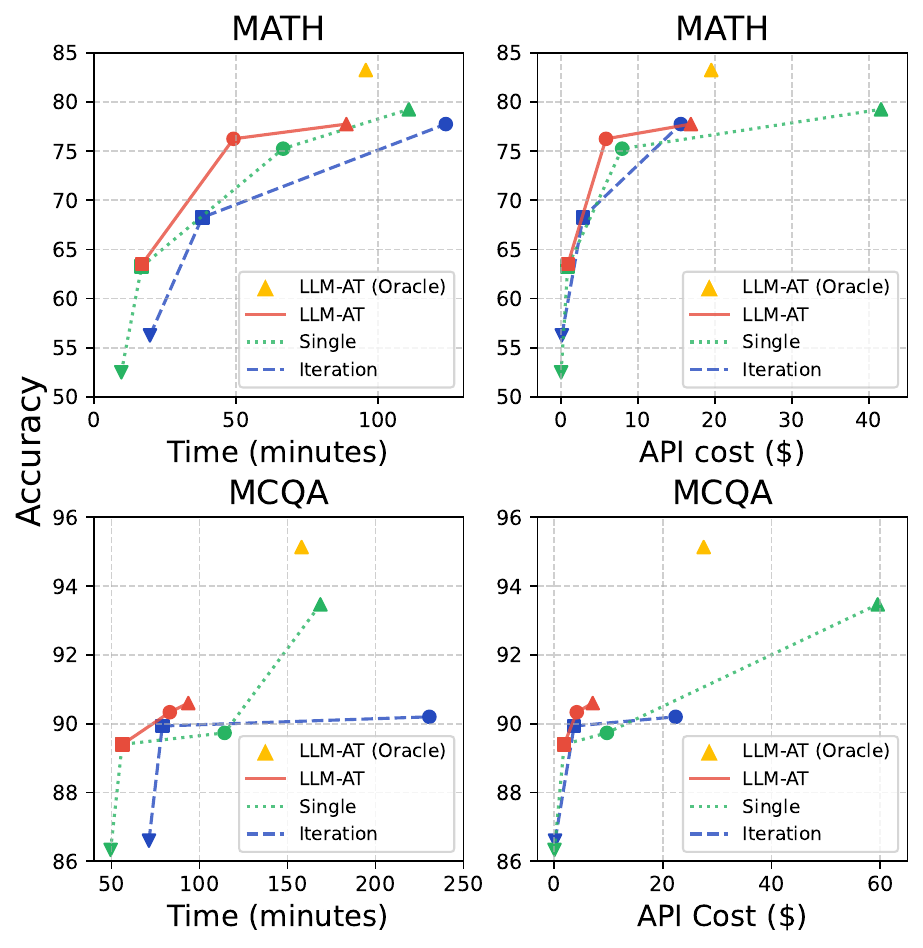}}
    \hspace{-1cm}
    \caption{Main results. 
    The marker shapes represent the LLM used (▲: o1, ●: o1-mini, ■: GPT-4o, ▼: GPT-4o-mini).
    For \ours(red), each marker indicates the top-tier model. \ours(Oracle) means the results obtained using an oracle judge.}
    \label{fig:main1}
\end{figure}

%Figure~\ref{fig:main1} shows the trade-offs between accuracy and latency, as well as accuracy and API cost. A curve positioned closer to the upper-left indicates that the method achieves higher performance while minimizing cost and latency. The curves of \ours consistently appear in the upper left of all MATH and MCQA graphs, compared to the single and iteration methods. This implies that \ours optimizes both accuracy-latency and accuracy-cost trade-offs.

%Specifically, \ours achieves substantial efficiency gains in terms of latency. For MATH, \ours reduces latency by 28.24\% (123.73 min $\rightarrow$ 88.79 min), compared to the o1-mini iteration, while preserving accuracy. \ours also attains a 19.81\% latency reduction (110.73.min $\rightarrow$ 88.79 min) and comparable performance to the single o1. For MCQA, \ours reduces latency by 59.34\% (230.54 min$\rightarrow$ 93.69 min) while outperforming the o1-mini iteration.

%Additionally, \ours demonstrates significant monetary cost efficiency. For MATH, our approach achieves a 59.37\% cost reduction (\$ 41.56 $\rightarrow$ \$ 16.89) compared to the single o1, while showing slightly lower performance. For MCQA, the single o1 exhibits the highest performance but requires 88.01\% more in API costs (\$ 59.52) than \ours (\$ 7.14). Therefore, \ours serves as the ideal choice in practical scenarios that demand both cost efficiency and high performance.

Figure~\ref{fig:main1} shows the trade-offs between accuracy and total execution time as well as accuracy and API cost.
A curve positioned closer to the upper-left indicates that the method achieves higher performance while minimizing the cost and total execution time.
The curves of \ours appear consistently in the upper left of all MATH and MCQA graphs, compared to the single and iteration methods.
This implies that \ours optimizes both the accuracy-time and accuracy-cost trade-offs.

Specifically, \ours achieves substantial time efficiency while maintaining comparable accuracy.
For MATH, \ours reduces time by 28.24\% (123.73 → 88.79 min) over the o1-mini iteration and by 19.81\% (110.73 → 88.79 min) compared to the single o1.
For MCQA, \ours reduces the time by 59.34\% (230.54 → 93.69 min) and outperforms the o1-mini iteration.
Furthermore, \ours also shows strong cost efficiency.
For MATH, \ours reduces cost by 59.37\% (\$41.56 → \$16.89) with slightly lower performance than the single o1.
For MCQA, while the single o1 performs the best, it costs 88.01\% more than \ours (\$59.52 vs. \$7.14).

% In particular, \suggestion{\ours achieves} higher accuracy with \suggestion{lower execution} time and cost than baselines.
%\suggestion{Even when restricted to the top-tier model, \ours achieves higher accuracy than the baselines with lower cost and execution time.}
Some results of \ours show a higher accuracy than the baselines with a lower cost and execution time.
\ours capped at o1 with the oracle judge achieves a more efficient performance–cost trade-off than the single o1.
Likewise, \ours capped at o1-mini outperforms both the single and iteration baselines with o1-mini.
These results indicate that \ours enables more effective inference by adaptively selecting model tiers based on the input question.
Further analysis is provided in \S~\ref{sec:reversal} to support these findings.
% A detailed analysis of these results is provided in \S~\ref{sec:reversal}.

\subsection{Evaluating the \oursacc} 
%\subsection{Distribution of the Estimated Accuracy}
%\todo{appendix ablation 결과 언급 }
\begin{figure}[t]
    \centering
    \includegraphics[width=\columnwidth]{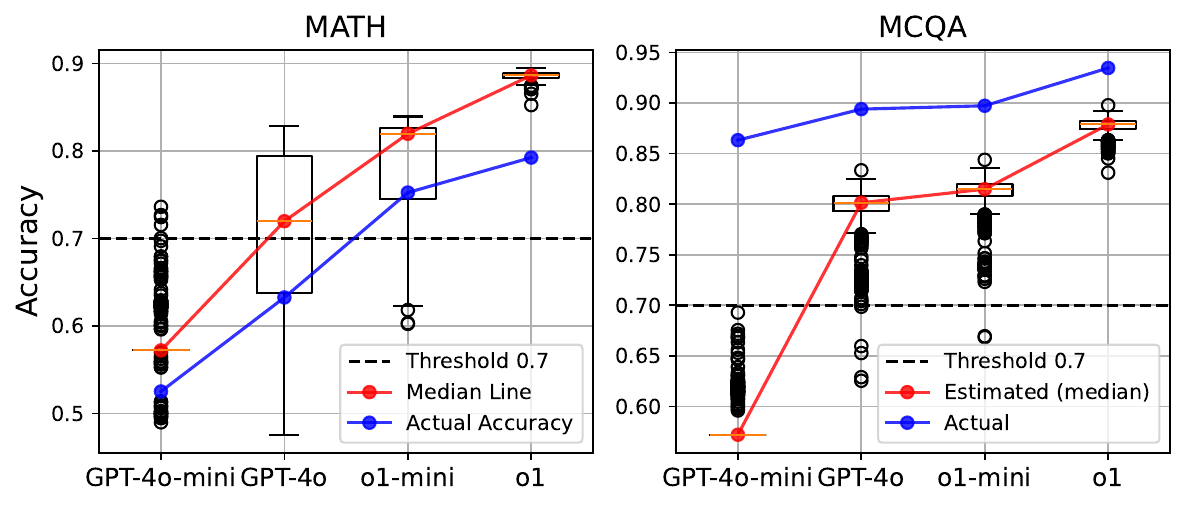}
     \caption{Distribution of the estimated accuracy.} 
    \label{fig:prob}
\end{figure}

To validate the effectiveness of our \oursacc, we compare the estimated accuracy with the actual accuracy obtained through single inferences from each LLM tier.
Figure~\ref{fig:prob} shows the distribution of the estimated accuracy along with the actual accuracy.
For MATH and MCQA, the median of the estimated accuracy distribution for each tier aligns well with the trend of the actual accuracy. 
This result demonstrates that, even without a correctness label, it is possible to estimate the accuracy of each tier for a given question and effectively model the relative performance differences between tiers using historical inference data.
In addition, despite incorporating other benchmark scores as smoothing factors, the close alignment between estimated and actual accuracy suggests that \oursacc can be applied flexibly to new datasets and various tier systems.

\begin{table}[t]
\centering
\resizebox{\columnwidth}{!}{%
\begin{tabular}{llccc}
\toprule
 &\textbf{Quantile} & \textbf{Accuracy} & \makecell{\textbf{API Cost}\\\textbf{(\$)}} & \makecell{\textbf{Time}\\\textbf{(minutes)}} \\
\midrule
\multirow{4}{*}{\rotatebox{90}{\textbf{MATH}}} 
& Q1 ($q_1$-- $q_{100}$)& 0.71 & 4.41 & 25.28 \\
& Q2 ($q_{101}$-- $q_{200}$)& 0.79 & 4.81 & 23.70 \\
& Q3 ($q_{201}$-- $q_{300}$)& 0.75 & 3.73 & 20.44 \\
& Q4 ($q_{301}$-- $q_{400}$)& 0.86 & 3.94 & 19.37 \\
\midrule
\multirow{4}{*}{\rotatebox{90}{\textbf{MCQA}}} 
& Q1 ($q_1$-- $q_{375}$)& 0.835 & 2.46 & 25.28 \\
& Q2 ($q_{376}$-- $q_{750}$)& 0.893 & 2.25 & 23.70 \\
& Q3 ($q_{751}$-- $q_{1,125}$)& 0.941 & 1.11 & 20.44 \\
& Q4 ($q_{1,126}$-- $q_{1,500}$)& 0.955 & 1.32 & 19.37 \\
\bottomrule
\end{tabular}
}
\caption{Performance by the amount of accumulated historical data. Q1 corresponds to the lowest 25\%, Q2 to the 25-50\% range, Q3 to the 50-75\% range, and Q4 includes questions in the top 25\% of entire dataset.
}
\label{tab:quartile_performance}
\end{table}

\begin{table*}[t]
\centering
\resizebox{\linewidth}{!}{%
\begin{tabular}{lccc|ccc}
\toprule
\multirow{1}{*}{} & \multicolumn{3}{c}{\textbf{MATH}} & \multicolumn{3}{c}{\textbf{MCQA}} \\
\cmidrule(lr){2-4} \cmidrule(lr){5-7}
\textbf{Setting} & \textbf{Accuracy} & \textbf{API Cost (\$) } & \textbf{Time (minutes)} & \textbf{Accuracy} & \textbf{API Cost (\$)} & \textbf{Time (minutes)} \\
\midrule
Recent 30 history  & 0.770 & 15.33 & 90.97 & 0.911 & 7.89  & 97.20 \\
Full accumulation  & 0.778 & 16.89 & 88.79 & 0.906 & 7.14  & 93.69 \\
\bottomrule
\end{tabular}
}
\caption{Performance based on the quality of the 30 most recent historical data. Recent 30 history setting constructs the history using only the 30 most recent past entries for each input. Full accumulation refers to the original setting, where the history is built by sequentially accumulating all past data.}
\label{tab:recent_vs_main}
\end{table*}

\subsection{Performance of the Judge}

The performance of the judge affects not only the generation of pseudo-labels for each tier but also the overall performance of \ours.
Table~\ref{tab:judge performance} compares the F1, recall, and precision of the judge with the accuracy of the generator to assess the reliability of the judge.
The judge shows reliable F1 scores on both MATH and MCQA. In particular, in both datasets, the F1 score of the judge is higher than the accuracy of the generator, especially on the more challenging MATH task.
In addition, using the low-performing GPT-4o-mini as a judge leads to degraded performance; this is mitigated by a special judge, which helps compensate for the limited self-verification ability of the weaker model.

Additionally, the correct and incorrect pseudo-labels for each model are determined by the evaluation of the judge.
Based on the reliable performance of the judge, we consider that the pseudo-labels of correctness can be trustworthy, and believe that they support the robustness of \ours.
Furthermore, this result suggests that ours can work effectively even in real-world scenarios where gold labels or human feedback are not available.

\begin{table*}[t]
\centering
\resizebox{\linewidth}{!}{%
\begin{tabular}{llccccccc}
\toprule
 & & \textbf{Geometry} & \textbf{Algebra} & \makecell{\textbf{Intermediate}\\\textbf{Algebra}} & \textbf{Prealgebra} & \textbf{Precalculus} & \makecell{\textbf{Counting}\\\textbf{\& Probability}} & \makecell{\textbf{Number}\\\textbf{Count}} \\
\midrule
\multirow{3}{*}{{\textit{Accuracy}}}
% \multirow{3}{*}{\rotatebox{90}{\textit{Accuracy}}} 
  & GPT-4o   & 0.421 & 0.732 & 0.604 & 0.726 & 0.386 & 0.807 & \textbf{0.813} \\
  & o1-mini  & \textbf{0.667} & \textbf{0.875} & \textbf{0.729} & \textbf{0.843} & \textbf{0.474} & \textbf{0.895} & 0.781 \\
  &\textit{Diff} & \cellcolor{green7}0.246 & \cellcolor{green6}0.143 & \cellcolor{green5}0.125 & \cellcolor{green4}0.117 & \cellcolor{green3}0.088  & \cellcolor{green2}0.088  & \cellcolor{green1}-0.032 \\
\midrule
\multirow{2}{*}{\textit{Proportion}} 
% {\makecell{\textit{Pro-}\\\textit{portion}}}
  & GPT-4o   & 0.333 & \underline{0.482} & 0.281 & \underline{0.549} & \underline{0.750} & \underline{0.328} & \underline{0.526} \\
  & o1-mini  & \underline{0.491} & \underline{0.482} & \underline{0.333} & 0.373 & 0.156 & \underline{0.328} & 0.456 \\
\bottomrule
\end{tabular}%
}
\caption{Comparison of accuracy and final-tier selection proportions between GPT-4o and o1-mini in MATH.}
\label{tab:performance reversals}
\end{table*}

\subsection{Robustness to History Quality and Cold Start Problem}
% \subsection{Robustness to \suggestion{the Cold Start Problem}}
The starter utilizes the history to estimate the accuracy of each tier for a given question.
Therefore, performance can be affected by the amount and quality of historical data, especially in cold-start scenarios.
To investigate this effect, we examine how performance varies with the amount and quality of historical data.
Table~\ref{tab:quartile_performance} shows the change in performance of \ours in the quartiles of the accumulated historical data.
In both MATH and MCQA, performance is relatively low in the first quartile due to cold-start effects, but improves significantly from the second quartile.
This indicates that just a few hundred examples are sufficient to improve performance---a number that can be quickly reached in the real-world LLM service.

Table~\ref{tab:recent_vs_main} shows the performance when only the 30 most recent questions are stored for each input question.
Even when historical data changes with every input, \ours maintains a performance comparable to the original results.
This demonstrates that the framework remains effective even when the history database contains limited or inconsistent quality.
These results indicate that \ours is robust to both the size and quality of the history and maintains stable performance even under a cold start scenario.

\subsection{Robustness to Performance Reversals}
\label{sec:reversal}
Although the LLM performance generally follows the order of the tiers, there are cases where the lower-tier models outperform the higher-tier for specific questions~\cite{ding2024hybrid, aggarwal2024automix}. 
To evaluate the robustness of \ours under such conditions, we analyze the selection results of tiers in MATH.
Table~\ref{tab:performance reversals} shows the difference in accuracy between GPT-4o and o1-mini in seven categories of MATH, along with the proportion of the two models selected as the final tier in \ours. 
Toward the right side of the table, the accuracy difference narrows, and in the Number Theory, GPT-4o even outperforms o1-mini.  
Reflecting this, GPT-4o is selected more frequently as the performance gap becomes smaller.
This selection tendency appears to result from the starter strategy of choosing the lowest tier among those whose estimated accuracy exceeds a certain threshold. 
These results show that \ours remains robust even in cases of performance reversals across model tiers, which also explains why some results outperform the single baselines.

\subsection{System Overhead Analysis}

\begin{table}[t]
\centering
\resizebox{\columnwidth}{!}{%
\begin{tabular}{lcc|cc}
\toprule
\multirow{2}{*}{} & \multicolumn{2}{c|}{\textbf{MATH}} & \multicolumn{2}{c}{\textbf{MCQA}} \\
\cmidrule(lr){2-3} \cmidrule(lr){4-5}
 \textbf{Module} & \makecell{\textbf{API Cost}\\\textbf{(\$)}} & \makecell{\textbf{Time}\\\textbf{(minutes)}} & \makecell{\textbf{API Cost}\\\textbf{(\$)}} & \makecell{\textbf{Time}\\\textbf{(minutes)}} \\
\midrule
Generator & 13.20 & 62.97 & 4.79 & 66.92 \\
Judge     & 3.69  & 25.82 & 2.35 & 26.77 \\
Starter   & -     & 0.50   & -    & 1.50 \\
\bottomrule
\end{tabular}
}
\caption{Execution time and API cost overhead by module.}
\label{tab:overhead}
\end{table}

\begin{table}[h]
\centering
\resizebox{0.68\columnwidth}{!}{%
\begin{tabular}{@{}crr@{}} 
\toprule
\textbf{\# of transition} & \textbf{MATH}  & \textbf{MCQA}\\
\midrule
0 & 83.5\% & 95.9\% \\
1 & 15.2\% & 3.3\%  \\
2 & 1.3\% & 0.8\%  \\
3 & 0.0\% & 0.0\% \\
\bottomrule
\end{tabular}
}
\caption{Distribution of tier transition counts.}
\label{tab:transition}
\end{table}

If the time and cost of the auxiliary modules are excessively high compared to the generator, which is central to the generation of LLM, the overall system efficiency may be compromised.
Moreover, if \ours requires more time and cost than using a single LLM, the benefits of the hierarchical approach would be diminished, making it inefficient. Therefore, we analyze the overhead of \ours in terms of execution time, cost, and the number of tier transitions.

\minisection{Time and Cost}
Table~\ref{tab:overhead} shows the execution time and API cost for each module. 
The judge and starter introduce minimal overhead compared to the generator.
%In terms of time, the starter operates approximately 44.6 to 125 times faster than the generator, and the judge is also about 2.5 times faster. 
The starter is approximately 44.6 to 125 times faster, and the judge is about 2.5 times faster than the generator.
In terms of cost, the judge incurs 2 to 4 times less than the generator, while the starter generates none, as it uses an open-source embedding model.
Given that the embedding model is much smaller than the generation model, its computational overhead is also expected to be significantly lower than that of the generator.

\minisection{Tier Transition}
Table~\ref{tab:transition} shows the distribution of the tier transition counts for each dataset. 
In both, the majority of questions are resolved with a single inference, which means that the final answer is obtained from the initially selected tier without any transitions.
As discussed in \S~\ref{sec:main}, \ours shows a more cost-effective performance compared to using a single LLM. 
This suggests that the starter effectively reduces the number of tier transitions by selecting an appropriate initial tier in most cases, thereby mitigating the cumulative overhead that can arise in a hierarchical structure.

\subsection{Tier Selection Depending on Question Difficulty} % task, question, question

\begin{figure}[t]
    \centering
    \includegraphics[width=\columnwidth]{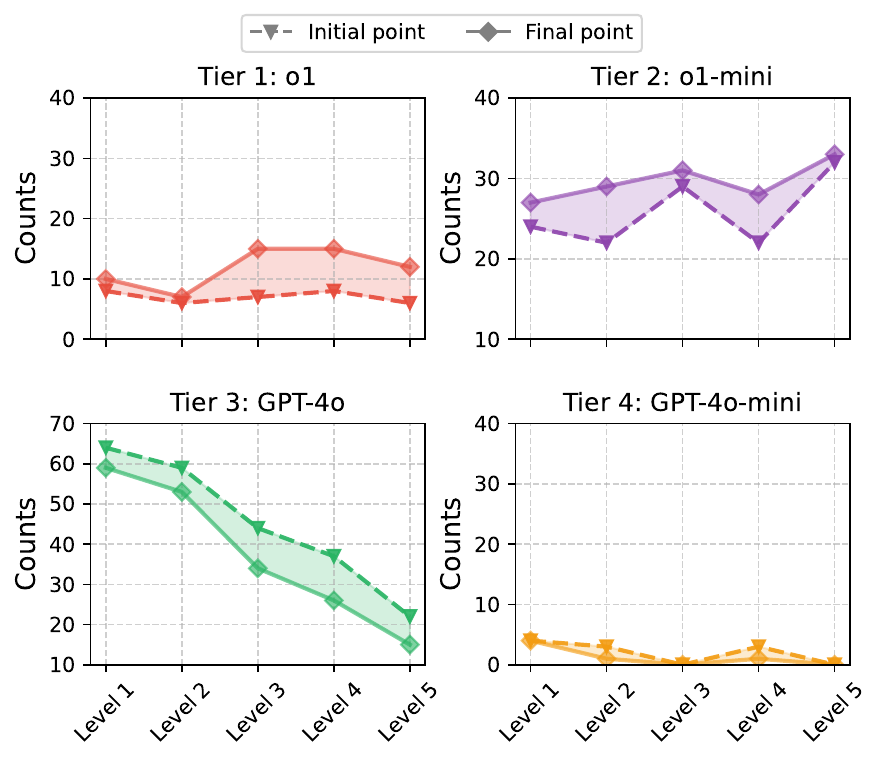}
    \caption{Tiers selected by \ours in MATH based on question difficulty.} 
    \label{fig:main2_math}
\end{figure}

To optimize the trade-off between cost and performance, it is ideal to solve low-difficulty problems using lower-tier models with lower costs while addressing high-difficulty problems with higher-tier models that are more expensive but offer superior performance.
Figure~\ref{fig:main2_math} shows how \ours selects the tiers in this ideal way.
The relatively lower-tier model, GPT-4o, is selected less frequently as the difficulty of the question increases. 
Conversely, the higher-tier models, o1-mini and o1, are selected more frequently as the question difficulty rises.
Additionally, lower-tier models are selected more frequently as initial points than as final points, while the opposite is observed for higher-tier models.
This supports our proposed method of selecting an appropriate initial tier and, when necessary, transitioning to a higher-performing tier for the final answer, thereby optimizing the performance efficiency relative to cost.

\section{Conclusion}

We introduce \ours, a novel framework that automatically selects LLM tiers to maximize performance while reducing cost and execution time.
We also propose \oursacc, which estimates the accuracy of each LLM tier based on past inference records without answer labels, enabling the selection of the most appropriate starting tier expected to solve a given question.
Our approach requires no additional training or supervised fine-tuning, making it readily applicable in real-world scenarios.
We conduct extensive experiments on datasets ranging from simple to graduate-level questions, demonstrating that \ours optimizes the trade-offs between accuracy and cost as well as accuracy and execution time.

\section*{Limitations}

In this study, we focus on QA tasks with clearly defined answers. As a future direction, it would be important to extend the \ours system to open-ended generation tasks.
In addition, integrating the heterogeneous tier systems of different LLM providers such as OpenAI, Google, and Anthropic remains a challenging task. Therefore, our experiments are conducted solely based on OpenAI’s tier system, which is widely used in the NLP research community. 
Future work may explore unified tier systems that integrate both open-source and proprietary LLMs to further optimize cost-performance trade-offs.

% In addition, while \oursacc effectively captures performance trends across different LLMs, there is still a slight error between estimated and actual performance.
%This could be mitigated by incorporating pseudo-labeling with ground truth or human feedback.

\section*{Acknowledgements}
% RS-2025_02264000
This work was supported by the National R\&D Program for Cancer Control through the National Cancer Center(NCC) funded by the Ministry of Health \& Welfare, Republic of Korea (No.RS-2025-02264000).
This work was supported by Institute of Information \& communications Technology Planning \& Evaluation(IITP) grant funded by the Korea government(MSIT) (No.RS-2023-00261068, Development of Lightweight Multimodal AntiPhishing Models and Split-Learning Techniques for Privacy-Preserving Anti-Phishing).

% 다양한 티어 시스템을 한 번에 활용하는 경우도 있는데 그런거는 고려하지 않았따. Claude, Gemini를 같이 쓴다던가.
% Opensource를 포함해서 LLM Provider들의 Heterogeneous한 티어 시스템 (Claude, Gemini)을 통합하는 것을 고려하지 않았다.

%Since integrating heterogeneous tier systems across different LLM providers, such as OpenAI, Google, and Anthropic, is hard, we employ OpenAI's tier system, which is widely studied in the NLP research community.
%A future direction would be optimizing the trade-offs between performance and inference costs using a unified tier system that integrates both open-source and proprietary LLMs.

% 서로 다른 LLM 제공 업체들의 LLM system들 간의 일관된 티어 시스템이 존재하는 것을 어렵기 때문에,
% 예를 본 연구에서는 기존 연구들에서 많이 활용되는 OpenAI의 티어 시스템을 LLM-AT에서 활용하였다. 
% 향후 연구에서는 opensource model을 포함하여 이질적인 시스템을 통합하여 LLM-AT를 적용해볼 수 있을 것이다.

\bibliography{ref}

\begin{thebibliography}{29}
\providecommand{\natexlab}[1]{#1}

\bibitem[{Aggarwal et~al.(2024)Aggarwal, Madaan, Anand, Potharaju, Mishra, Zhou, Gupta, Rajagopal, Kappaganthu, Yang et~al.}]{aggarwal2024automix}
Pranjal Aggarwal, Aman Madaan, Ankit Anand, Srividya~Pranavi Potharaju, Swaroop Mishra, Pei Zhou, Aditya Gupta, Dheeraj Rajagopal, Karthik Kappaganthu, Yiming Yang, et~al. 2024.
\newblock Automix: Automatically mixing language models.
\newblock \emph{Advances in Neural Information Processing Systems}, 37:131000--131034.

\bibitem[{Anthropic(2023)}]{Claude}
Anthropic. 2023.
\newblock \href {https://www.anthropic.com/claude} {Meet claude}.
\newblock \url{https://www.anthropic.com/claude}.

\bibitem[{Chen et~al.(2024{\natexlab{a}})Chen, Zaharia, and Zou}]{chen2024frugalgpt}
Lingjiao Chen, Matei Zaharia, and James Zou. 2024{\natexlab{a}}.
\newblock \href {https://openreview.net/forum?id=cSimKw5p6R} {Frugal{GPT}: How to use large language models while reducing cost and improving performance}.
\newblock \emph{Transactions on Machine Learning Research}.

\bibitem[{Chen et~al.(2024{\natexlab{b}})Chen, Jiang, Lin, Kwok, and Zhang}]{chen2024routerdc}
Shuhao Chen, Weisen Jiang, Baijiong Lin, James Kwok, and Yu~Zhang. 2024{\natexlab{b}}.
\newblock \href {https://openreview.net/forum?id=7RQvjayHrM} {Router{DC}: Query-based router by dual contrastive learning for assembling large language models}.
\newblock In \emph{The Thirty-eighth Annual Conference on Neural Information Processing Systems}.

\bibitem[{Chen et~al.(2023)Chen, Ma, Wang, and Cohen}]{chen2022program}
Wenhu Chen, Xueguang Ma, Xinyi Wang, and William~W. Cohen. 2023.
\newblock Program of thoughts prompting: Disentangling computation from reasoning for numerical reasoning tasks.
\newblock \emph{Transactions on Machine Learning Research}.

\bibitem[{Clark et~al.(2018)Clark, Cowhey, Etzioni, Khot, Sabharwal, Schoenick, and Tafjord}]{allenai:arc}
Peter Clark, Isaac Cowhey, Oren Etzioni, Tushar Khot, Ashish Sabharwal, Carissa Schoenick, and Oyvind Tafjord. 2018.
\newblock Think you have solved question answering? try arc, the ai2 reasoning challenge.
\newblock \emph{arXiv:1803.05457v1}.

\bibitem[{DeepSeek(2025)}]{DeepSeek}
DeepSeek. 2025.
\newblock \href {https://www.deepseek.com/} {Deepseek into the unknown}.
\newblock \url{https://www.deepseek.com/}.

\bibitem[{Ding et~al.(2024)Ding, Mallick, Wang, Sim, Mukherjee, R{\"u}hle, Lakshmanan, and Awadallah}]{ding2024hybrid}
Dujian Ding, Ankur Mallick, Chi Wang, Robert Sim, Subhabrata Mukherjee, Victor R{\"u}hle, Laks V.~S. Lakshmanan, and Ahmed~Hassan Awadallah. 2024.
\newblock \href {https://openreview.net/forum?id=02f3mUtqnM} {Hybrid {LLM}: Cost-efficient and quality-aware query routing}.
\newblock In \emph{The Twelfth International Conference on Learning Representations}.

\bibitem[{Feng et~al.(2024)Feng, Shi, Wang, Ding, Balachandran, and Tsvetkov}]{Abstain}
Shangbin Feng, Weijia Shi, Yike Wang, Wenxuan Ding, Vidhisha Balachandran, and Yulia Tsvetkov. 2024.
\newblock \href {https://aclanthology.org/2024.acl-long.786} {Don{'}t hallucinate, abstain: Identifying {LLM} knowledge gaps via multi-{LLM} collaboration}.
\newblock In \emph{Proceedings of the 62nd Annual Meeting of the Association for Computational Linguistics (Volume 1: Long Papers)}, pages 14664--14690, Bangkok, Thailand. Association for Computational Linguistics.

\bibitem[{Fu et~al.(2024)Fu, Qi, Gao, Jiang, Dong, and Zhou}]{fu-etal-2024-msi}
Dayuan Fu, Biqing Qi, Yihuai Gao, Che Jiang, Guanting Dong, and Bowen Zhou. 2024.
\newblock \href {https://doi.org/10.18653/v1/2024.emnlp-main.38} {{MSI}-agent: Incorporating multi-scale insight into embodied agents for superior planning and decision-making}.
\newblock In \emph{Proceedings of the 2024 Conference on Empirical Methods in Natural Language Processing}, pages 643--659, Miami, Florida, USA. Association for Computational Linguistics.

\bibitem[{Gou et~al.(2024)Gou, Shao, Gong, yelong shen, Yang, Duan, and Chen}]{gou2024critic}
Zhibin Gou, Zhihong Shao, Yeyun Gong, yelong shen, Yujiu Yang, Nan Duan, and Weizhu Chen. 2024.
\newblock \href {https://openreview.net/forum?id=Sx038qxjek} {{CRITIC}: Large language models can self-correct with tool-interactive critiquing}.
\newblock In \emph{The Twelfth International Conference on Learning Representations}.

\bibitem[{Hendrycks et~al.(2021{\natexlab{a}})Hendrycks, Burns, Basart, Critch, Li, Song, and Steinhardt}]{mmlu}
Dan Hendrycks, Collin Burns, Steven Basart, Andrew Critch, Jerry Li, Dawn Song, and Jacob Steinhardt. 2021{\natexlab{a}}.
\newblock Aligning ai with shared human values.
\newblock \emph{Proceedings of the International Conference on Learning Representations (ICLR)}.

\bibitem[{Hendrycks et~al.(2021{\natexlab{b}})Hendrycks, Burns, Kadavath, Arora, Basart, Tang, Song, and Steinhardt}]{hendrycksmath2021}
Dan Hendrycks, Collin Burns, Saurav Kadavath, Akul Arora, Steven Basart, Eric Tang, Dawn Song, and Jacob Steinhardt. 2021{\natexlab{b}}.
\newblock Measuring mathematical problem solving with the math dataset.
\newblock \emph{NeurIPS}.

\bibitem[{Huang et~al.(2023)Huang, Chen, Mishra, Zheng, Yu, Song, and Zhou}]{huang2023large}
Jie Huang, Xinyun Chen, Swaroop Mishra, Huaixiu~Steven Zheng, Adams~Wei Yu, Xinying Song, and Denny Zhou. 2023.
\newblock Large language models cannot self-correct reasoning yet.
\newblock \emph{arXiv preprint arXiv:2310.01798}.

\bibitem[{Jung et~al.(2019)Jung, Kim, and Shim}]{jung2019crowdsourced}
Woohwan Jung, Younghoon Kim, and Kyuseok Shim. 2019.
\newblock Crowdsourced truth discovery in the presence of hierarchies for knowledge fusion.
\newblock \emph{Advances in Database Technology-EDBT 2019}, pages 205--216.

\bibitem[{Lan et~al.(2024)Lan, Hu, Wang, Wang, Ye, Zhao, Lim, Xiong, and Wang}]{lan-etal-2024-llm}
Yihuai Lan, Zhiqiang Hu, Lei Wang, Yang Wang, Deheng Ye, Peilin Zhao, Ee-Peng Lim, Hui Xiong, and Hao Wang. 2024.
\newblock \href {https://doi.org/10.18653/v1/2024.emnlp-main.7} {{LLM}-based agent society investigation: Collaboration and confrontation in avalon gameplay}.
\newblock In \emph{Proceedings of the 2024 Conference on Empirical Methods in Natural Language Processing}, pages 128--145, Miami, Florida, USA. Association for Computational Linguistics.

\bibitem[{Lu et~al.(2024)Lu, Yuan, Lin, Lin, Yuan, Zhou, and Zhou}]{lu-etal-2024-routing}
Keming Lu, Hongyi Yuan, Runji Lin, Junyang Lin, Zheng Yuan, Chang Zhou, and Jingren Zhou. 2024.
\newblock \href {https://doi.org/10.18653/v1/2024.naacl-long.109} {Routing to the expert: Efficient reward-guided ensemble of large language models}.
\newblock In \emph{Proceedings of the 2024 Conference of the North American Chapter of the Association for Computational Linguistics: Human Language Technologies (Volume 1: Long Papers)}, pages 1964--1974, Mexico City, Mexico. Association for Computational Linguistics.

\bibitem[{Madaan et~al.(2023)Madaan, Tandon, Gupta, Hallinan, Gao, Wiegreffe, Alon, Dziri, Prabhumoye, Yang, Gupta, Majumder, Hermann, Welleck, Yazdanbakhsh, and Clark}]{madaan2023selfrefine}
Aman Madaan, Niket Tandon, Prakhar Gupta, Skyler Hallinan, Luyu Gao, Sarah Wiegreffe, Uri Alon, Nouha Dziri, Shrimai Prabhumoye, Yiming Yang, Shashank Gupta, Bodhisattwa~Prasad Majumder, Katherine Hermann, Sean Welleck, Amir Yazdanbakhsh, and Peter Clark. 2023.
\newblock \href {https://openreview.net/forum?id=S37hOerQLB} {Self-refine: Iterative refinement with self-feedback}.
\newblock In \emph{Thirty-seventh Conference on Neural Information Processing Systems}.

\bibitem[{Maurya et~al.(2024)Maurya, Srivatsa, and Kochmar}]{maurya2024selectllm}
Kaushal~Kumar Maurya, KV~Srivatsa, and Ekaterina Kochmar. 2024.
\newblock Selectllm: Query-aware efficient selection algorithm for large language models.
\newblock \emph{arXiv preprint arXiv:2408.08545}.

\bibitem[{Mihaylov et~al.(2018)Mihaylov, Clark, Khot, and Sabharwal}]{OpenBook}
Todor Mihaylov, Peter Clark, Tushar Khot, and Ashish Sabharwal. 2018.
\newblock \href {https://doi.org/10.18653/v1/D18-1260} {Can a suit of armor conduct electricity? a new dataset for open book question answering}.
\newblock In \emph{Proceedings of the 2018 Conference on Empirical Methods in Natural Language Processing}, pages 2381--2391, Brussels, Belgium. Association for Computational Linguistics.

\bibitem[{Ong et~al.(2024)Ong, Almahairi, Wu, Chiang, Wu, Gonzalez, Kadous, and Stoica}]{ong2024routellm}
Isaac Ong, Amjad Almahairi, Vincent Wu, Wei-Lin Chiang, Tianhao Wu, Joseph~E Gonzalez, M~Waleed Kadous, and Ion Stoica. 2024.
\newblock Routellm: Learning to route llms with preference data.
\newblock \emph{arXiv preprint arXiv:2406.18665}.

\bibitem[{OpenAI(2022)}]{Chatgpt}
OpenAI. 2022.
\newblock \href {https://openai.com/chatgpt/overview/} {Chatgpt. get answers. find inspiration. be more productive.}

\bibitem[{OpenAI(2025)}]{reasoningmodel}
OpenAI. 2025.
\newblock \href {https://platform.openai.com/docs/guides/reasoning/} {Reasoning models}.

\bibitem[{Rein et~al.(2024)Rein, Hou, Stickland, Petty, Pang, Dirani, Michael, and Bowman}]{rein2024gpqa}
David Rein, Betty~Li Hou, Asa~Cooper Stickland, Jackson Petty, Richard~Yuanzhe Pang, Julien Dirani, Julian Michael, and Samuel~R. Bowman. 2024.
\newblock \href {https://openreview.net/forum?id=Ti67584b98} {{GPQA}: A graduate-level google-proof q\&a benchmark}.
\newblock In \emph{First Conference on Language Modeling}.

\bibitem[{Wang et~al.(2024)Wang, Ma, Zhang, Ni, Chandra, Guo, Ren, Arulraj, He, Jiang et~al.}]{wang2024mmlupro}
Yubo Wang, Xueguang Ma, Ge~Zhang, Yuansheng Ni, Abhranil Chandra, Shiguang Guo, Weiming Ren, Aaran Arulraj, Xuan He, Ziyan Jiang, et~al. 2024.
\newblock Mmlu-pro: A more robust and challenging multi-task language understanding benchmark.
\newblock \emph{arXiv preprint arXiv:2406.01574}.

\bibitem[{Wei et~al.(2022)Wei, Wang, Schuurmans, Bosma, ichter, Xia, Chi, Le, and Zhou}]{NEURIPS2022_9d560961}
Jason Wei, Xuezhi Wang, Dale Schuurmans, Maarten Bosma, brian ichter, Fei Xia, Ed~Chi, Quoc~V Le, and Denny Zhou. 2022.
\newblock \href {https://proceedings.neurips.cc/paper_files/paper/2022/file/9d5609613524ecf4f15af0f7b31abca4-Paper-Conference.pdf} {Chain-of-thought prompting elicits reasoning in large language models}.
\newblock In \emph{Advances in Neural Information Processing Systems}, volume~35, pages 24824--24837. Curran Associates, Inc.

\bibitem[{Yao et~al.(2024)Yao, Yu, Zhao, Shafran, Griffiths, Cao, and Narasimhan}]{yao2024tree}
Shunyu Yao, Dian Yu, Jeffrey Zhao, Izhak Shafran, Tom Griffiths, Yuan Cao, and Karthik Narasimhan. 2024.
\newblock Tree of thoughts: Deliberate problem solving with large language models.
\newblock \emph{Advances in Neural Information Processing Systems}, 36.

\bibitem[{Yao et~al.(2023)Yao, Zhao, Yu, Du, Shafran, Narasimhan, and Cao}]{yao2023react}
Shunyu Yao, Jeffrey Zhao, Dian Yu, Nan Du, Izhak Shafran, Karthik~R Narasimhan, and Yuan Cao. 2023.
\newblock \href {https://openreview.net/forum?id=WE_vluYUL-X} {React: Synergizing reasoning and acting in language models}.
\newblock In \emph{The Eleventh International Conference on Learning Representations}.

\bibitem[{Zong et~al.(2024)Zong, Yan, Lu, Shao, Huang, Chang, and Zhuang}]{zong-etal-2024-triad}
Chang Zong, Yuchen Yan, Weiming Lu, Jian Shao, Yongfeng Huang, Heng Chang, and Yueting Zhuang. 2024.
\newblock \href {https://doi.org/10.18653/v1/2024.emnlp-main.101} {Triad: A framework leveraging a multi-role {LLM}-based agent to solve knowledge base question answering}.
\newblock In \emph{Proceedings of the 2024 Conference on Empirical Methods in Natural Language Processing}, pages 1698--1710, Miami, Florida, USA. Association for Computational Linguistics.

\end{thebibliography}

\appendix
\section{Additional Experimental Settings}
\label{detail setting}

The maximum output token length is set to 300 tokens for GPT-4o-mini and GPT-4o. 
However, for o1-mini and o1, due to the presence of unseen reasoning tokens \cite{reasoningmodel}, it is difficult to set an optimal maximum output token length, so no such limit is imposed on these models.
When measuring the API cost, we consider the input token cost and the output token cost, including reasoning tokens, but do not account for cached input.

\section{Additional Experimental Results}
\label{appendix: additional result}

\subsection{Smoothing Factor Specialized for the Benchmarks}
\label{specailized factor}

\begin{figure}[t]
    \centering
    \includegraphics[width=\columnwidth]{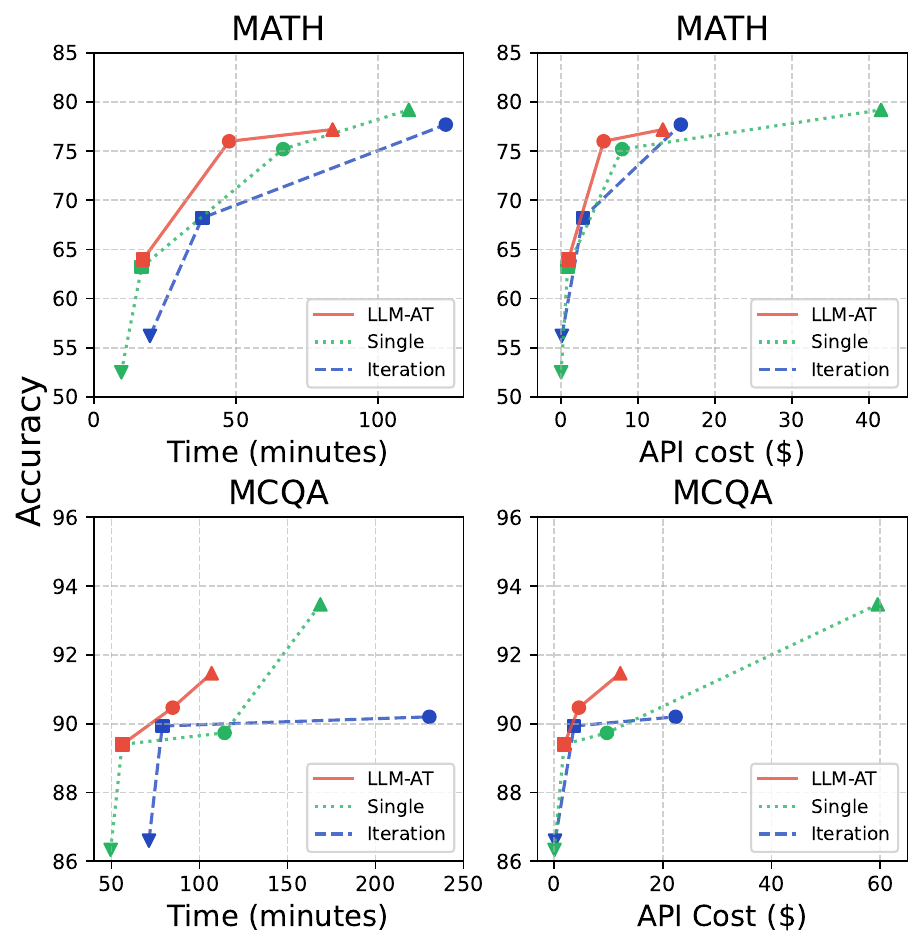}
    \caption{Results when using benchmark performance specialized for each benchmark as the $Acc^{Bench}$ value. For MATH, the official MATH performance was used, while for MCQA, the official MMLU performance was used.} 
    \label{fig:main_appendix}
\end{figure}

Figure~\ref{fig:main_appendix} presents the results when using benchmark performance similar to or identical to each dataset as $Acc^{Bench}$.
The results show that using benchmark performance similar to the target task as a smoothing factor can lead to additional performance improvements.
When MATH performance was used as a smoothing factor, the system maintained accuracy between the single baselines o1 and o1-mini, while achieving higher efficiency in time and cost.
Similarly, for MCQA, incorporating the MMLU performance resulted in better accuracy than o1-mini, along with improved efficiency in both time and cost.
In both cases, the results outperformed the main results obtained using the MMLU Pro accuracy as the smoothing factor.
These findings suggest that simply adjusting the smoothing factor without additional training allows \ours to be flexibly optimized for the given task.

\subsection{Comparison with Training-Based Approach}

We evaluated the performance of our training-free approach compared to RouteLLM~\cite{ong2024routellm}, a training-based routing method. RouteLLM uses a binary routing strategy that directs each input query to a strong or weak model. In our experiments, the top-tier model o1 is designated as the strong model and each lower-tier model is set as the weak model.

As shown in Tables~\ref{tab:routellm_MATH} and~\ref{tab:routellm_MCQA}, all the results of RouteLLM are very similar to the single inference with o1. 
This indicates that RouteLLM assigned most inputs to o1, ultimately failing to achieve an effective balance between accuracy and cost. 
In contrast, \ours achieves performance close to that of o1, but with significantly lower cost and execution time. 
These results demonstrate that training-based routing approaches rely on the model and data used to train the router, and their generalization performance can degrade significantly in out-of-distribution settings not seen during training.

\begin{table}[t]
\centering
\resizebox{\columnwidth}{!}{%
\begin{tabular}{lccc}
\toprule
\textbf{Method} & \textbf{Accuracy} & \makecell{\textbf{API Cost}\\\textbf{(\$)}} & \makecell{\textbf{Time}\\\textbf{(minutes)}}  \\
\midrule
RouteLLM$_{\mathrm{4o\text{-}mini}}$  & \textbf{0.795} & 41.08 & 109.85 \\
RouteLLM$_{\mathrm{4o}}$             & 0.790 & 40.99 & 109.22 \\
RouteLLM$_{\mathrm{o1\text{-}mini}}$ & 0.788 & 40.98 & 109.16 \\
o1$_{\mathrm{single}}$               & 0.793 & 41.56 & 110.73 \\
\rowcolor{gray!10}
LLM-AT                    & 0.778 & \textbf{16.89} & \textbf{88.79} \\
\bottomrule
\end{tabular}%
}
\caption{Performance comparison between training-based and training-free approaches on MATH.}
\label{tab:routellm_MATH}
\end{table}

\begin{table}[t]
\centering
\resizebox{\columnwidth}{!}{%
\begin{tabular}{lccc}
\toprule
\textbf{Method} & \textbf{Accuracy} & \makecell{\textbf{API Cost}\\\textbf{(\$)}} & \makecell{\textbf{Time}\\\textbf{(minutes)}}  \\
\midrule
RouteLLM$_{\mathrm{4o\text{-}mini}}$  & 0.929 & 53.14 & 162.28 \\
RouteLLM$_{\mathrm{4o}}$             & \textbf{0.935} & 51.60 & 151.99 \\
RouteLLM$_{\mathrm{o1\text{-}mini}}$ & 0.927 & 51.27 & 152.14 \\
o1$_{\mathrm{single}}$               & \textbf{0.935} & 59.52 & 168.68 \\
\rowcolor{gray!10}
LLM-AT                               & 0.906 & \textbf{16.89} & \textbf{93.69} \\
\bottomrule
\end{tabular}%
}
\caption{Performance comparison between training-based and training-free approaches on MCQA.}
\label{tab:routellm_MCQA} 
\end{table}

\subsection{Ablation Study}
\begin{figure}[t]
    \centering
    \includegraphics[width=\columnwidth]{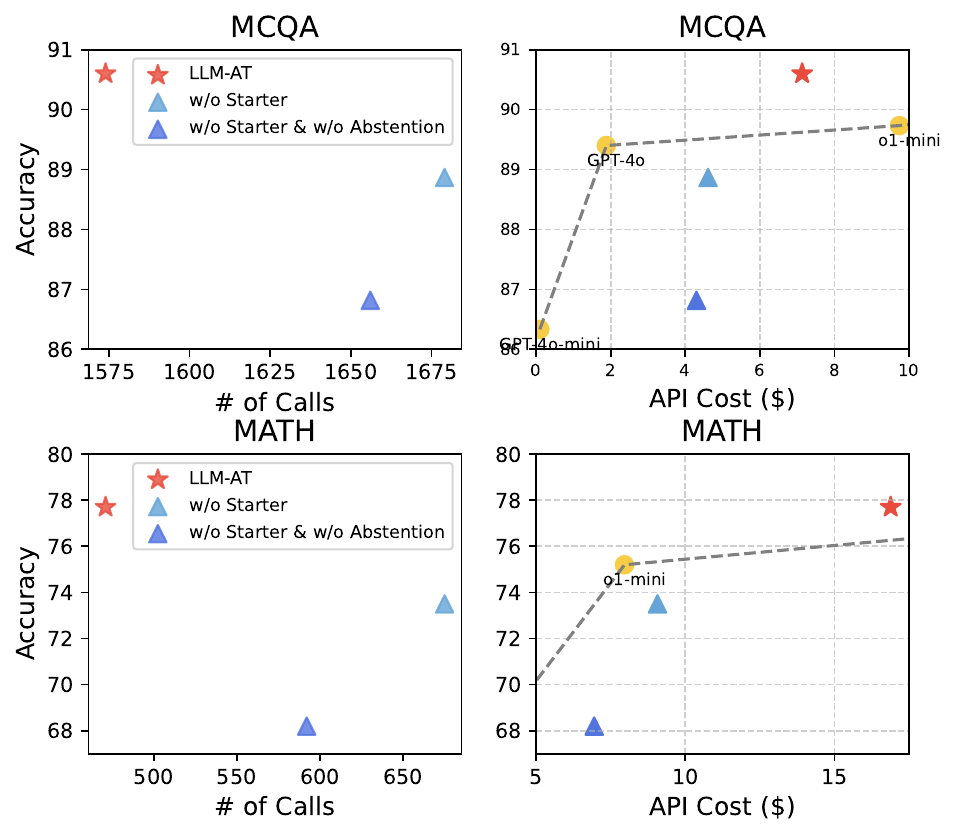}
    \caption{Ablation study. The left graph shows the accuracy and number of API calls, while the right graph presents the accuracy and API cost. In both graphs, points located toward the top-left indicate better performance.} 
    \label{fig:ablation}
\end{figure}

Figure~\ref{fig:ablation} presents the results of an ablation study conducted to analyze the effects of the starter module and the abstention option.
In the left plot, using the starter leads to improved performance across all datasets, while also significantly reducing the total number of API calls during inference. Although both \ours and the \texttt{w/o starter} use o1 as the top-tier model, the \texttt{w/o starter} appears inside the single inference baseline curve in the right plot. This indicates that it incurs higher costs while achieving lower performance compared to single-model inference. These results demonstrate that the starter effectively predicts the appropriate tier for each question, preventing performance degradation that may occur from underestimation when always starting from the lowest tier, and significantly enhancing cost efficiency and API call optimization.

Furthermore, providing the abstention option to lower-tier models can significantly improve performance without substantially increasing costs.
In the right plot, when the abstention option is enabled(\texttt{w/o Starter}), we observe that the cost increase relative to the accuracy gain compared to the \texttt{w/o Starter\& Abstention} is highly efficient.
For example, in MCQA, offering the abstention option increases the cost from \$4.3 to \$4.61 and accuracy improves from 86 to 88.86 resulting in a cost-to-performance improvement ratio of 9.06. 
This indicates that the model, by abstaining from difficult questions, avoids unnecessary generator output and judge evaluation, while improving overall performance by forwarding those questions to higher-tier models.

\subsection{Tier Selection Result for MCQA.}

\begin{figure}[t]
    \centering
    \includegraphics[width=\columnwidth]{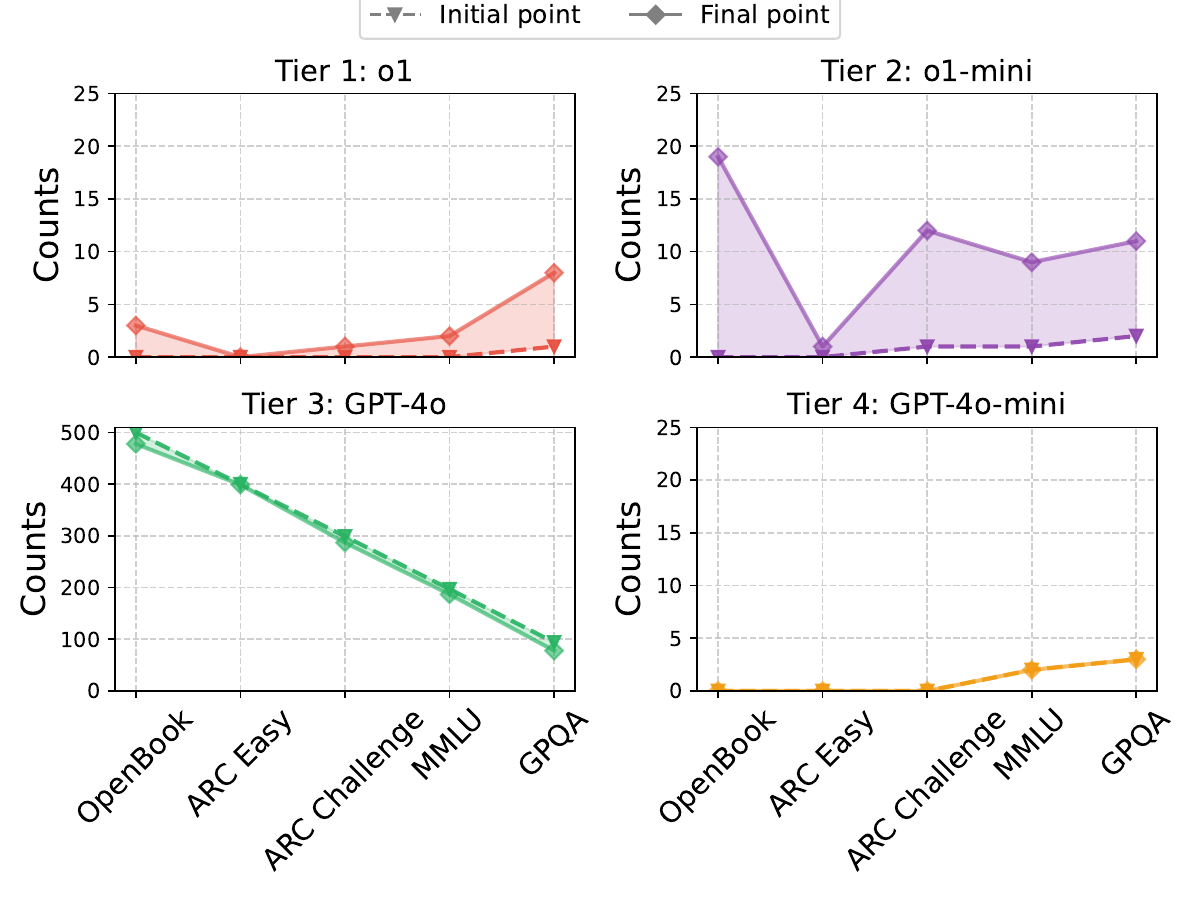}
    \caption{Tiers selected by \ours in MCQA based on question difficulty.} 
    \label{fig:main2_mcqa}
\end{figure}

As shown in the MCQA tier selection result in Figure~\ref{fig:main2_mcqa}, the selection frequency of the GPT-4o consistently decreases as the difficulty of the questions increases, while the selection frequency of o1-mini and o1 increases with higher level questions.
This demonstrates that \ours works consistently well not only for MATH but also for other tasks such as MCQA.
However, in MCQA, the performance gap between tiers is relatively smaller compared to MATH (Figure~\ref{fig:prob}), resulting in less noticeable differences in the selection frequency of each model as initial and final points.

\subsection{Abstention Analysis}
\begin{figure}[t]
    \centering
    \includegraphics[width=0.95\columnwidth]{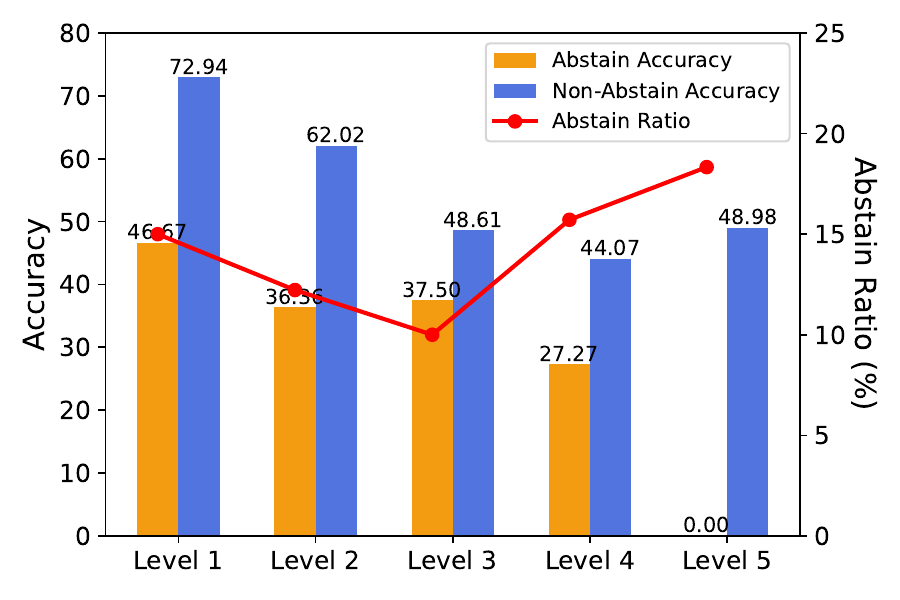}
    \caption{Performance of questions from abstained vs. non-abstained in MATH. The bar chart shows the accuracy on questions based on whether the model abstained or not, while the line chart shows the abstention rate for each difficulty level.} 
    \label{fig:abstain}
\end{figure}

Figure~\ref{fig:abstain} shows the analysis of whether a lower-tier model, when simply given an abstention option through the prompt, selectively abstains from difficult questions rather than doing so at random. To analyze this, we had the model re-answer the previously abstained questions and compared its performance with that on the questions it did not abstain from.

The analysis shows that, across all difficulty levels, the accuracy of the questions to which the model chose to abstain is consistently lower than the questions it attempted to answer. Furthermore, as the difficulty of the question increases, the abstention rate also increases and the accuracy on abstained questions gradually decreases. This indicates that the model is not abstaining at random, but rather selectively abstaining from questions that are more difficult overall or comparatively harder within the same difficulty level.

\section{Prompt Templates}
\label{appendix:prompts}
We provide the prompts used by the single and iteration baselines, as well as the prompts used by the generator and judge of \ours in Table~\ref{tab:prompt_COT} to ~\ref{tab:prompt_judge}.
For the judge, both \ours and the iteration baseline use the same prompt.
Following OpenAI's recommendation to avoid using CoT prompts for reasoning models, since they perform their own reasoning, we remove the sentence ‘Give step by step reasoning in a maximum of three sentences before your answer’ from the CoT prompt when using o1, the most complex reasoning model.

\begin{table*}[h]
    \centering
    \small
    \begin{tabular}{p{15cm}}
    \toprule
        Here is a question. An explanation is given before the final answer. 
        
        Answer the question yourself, giving your reasoning Be sure to generate the final answer in 'The correct answer is' format.
        \\
        \\
        Question: <question>\\
        Choices: <choices>
        \\
        \\
        The output format must strictly follow three inference statements and The correct answer is: (answer index (e.g., A) here). 
        Give step by step reasoning in a maximum of three sentences before your answer.
        \\Do not exceed three sentences. \\
    \bottomrule
    \end{tabular}
    \caption{
    % Chain-of-Thought prompt.
    Chain-of-Thought prompt for the single and iteration baselines and the \ours generator.
    }
    \label{tab:prompt_COT}
\end{table*}

\begin{table*}[h]
    \centering
    \small
    \begin{tabular}{p{15cm}}
    \toprule
        Review the following question and first determine whether you can solve it accurately.\\
        If you are confident that you can solve the problem, answer the question yourself, providing logical reasoning beforehand.\\
        Be sure to generate the final answer in 'The correct answer is' format.\\
        \\
        Question: <question>
        \\
        Choices: <choices>
        \\
        \\
        The output format must strictly follow three inference statements and The correct answer is: (answer index (e.g., A) here). Give step by step reasoning in a maximum of three sentences before your answer. 
        \\Do not exceed three sentences.\\
        \\
        If there is even the slightest doubt about solving the question correctly, output only “Abstain” and do not generate any additional content.\\ 
        Follow the instructions and handle the question appropriately without overestimating your abilities. \\
    \bottomrule
    \end{tabular}
    \caption{
    Chain-of-Thought prompt with an abstention option.
    % Chain-of-Thought prompt with an abstention option. In this study, only the lowest-tier model of \ours uses this prompt.
    }
    \label{tab:prompt_COT}
\end{table*}

\begin{table*}[h]
    \centering
    \small
    \begin{tabular}{p{15cm}}
    \toprule
        Given a mathematics question, write Python Code to solve the following questions. Store your result as a variable named ’answer’. \\
        The output format should follow the structure where ‘Code:’ is followed by the solution code, and ‘Answer:’ is followed by the answer. \\
        The final items you need to output are Code: code should be here\\
        \\
        Please output only the code after ‘Code:’ and do not generate any other content. \\
        If you have completed answering the question, do not generate unrelated content and end the response.\\
        \\
        Question: In how many ways can 4 books be selected from a shelf of 6 books if the order in which the books are selected does not matter?\\
        Code:\\
        %\begin{lstlisting}
        import math
        n = 6
        r = 4
        combinations = math.comb(n, r)
        answer = combinations
        %\end{lstlisting}
        \\
        \\
        Question: Find the distance between the points \$(2,1,-4)\$ and \$(5,8,-3)\$ \\
        Code: \\
        from sympy import sqrt \\
        x1, y1, z1 = 2, 1, -4 \\
        x2, y2, z2 = 5, 8, -3 \\
        distance\_squared = (x2 - x1)**2 + (y2 - y1)**2 + (z2 - z1)**2 \\
        distance = sqrt(distance\_squared) \\
        answer = distance \\
        \\
        Question: How many zeros are at the end of the product 25 \$ times\$ 240? \\
        Code: \\
        def count\_trailing\_zeros\_of\_product(num1, num2): \\
        	\hspace{4ex}product = num1 * num2  \\
        	\hspace{4ex}count\_zeros = 0 \\
        	\hspace{4ex}while product \% 10 == 0: \\
        		\hspace{4ex}\hspace{4ex}count\_\text{zeros} += 1 \\
        		\hspace{4ex}\hspace{4ex}product \//= 10 \\
        	\hspace{4ex}return count\_zeros \\
        num\_zeros = count\_trailing\_zeros\_of\_product(25, 240) \\
        answer = num\_zeros \\
        \\
        Question: <question>\\
        Code: \\
    \bottomrule
    \end{tabular}
    \caption{
    Program-of-Thought prompt for the single and iteration baselines and the \ours generator.}
    \label{tab:prompt_POT_baseline}
\end{table*}

\begin{table*}[h]
    \centering
    \small
    \begin{tabular}{p{15cm}}
    \toprule
        Given a mathematics question, write Python Code to solve the following questions. Store your result as a variable named ’answer’. \\
        The output format should follow the structure where ‘Code:’ is followed by the solution code, and ‘Answer:’ is followed by the answer. \\
        The final items you need to output are Code: code should be here\\
        \\
        Please output only the code after ‘Code:’ and do not generate any other content. \\
        If you have completed answering the question, do not generate unrelated content and end the response.\\
        \\
        Question: In how many ways can 4 books be selected from a shelf of 6 books if the order in which the books are selected does not matter?\\
        Code:\\
        %\begin{lstlisting}
        import math
        n = 6
        r = 4
        combinations = math.comb(n, r)
        answer = combinations
        %\end{lstlisting}
        \\
        \\
        Question: Find the distance between the points \$(2,1,-4)\$ and \$(5,8,-3)\$ \\
        Code: \\
        from sympy import sqrt \\
        x1, y1, z1 = 2, 1, -4 \\
        x2, y2, z2 = 5, 8, -3 \\
        distance\_squared = (x2 - x1)**2 + (y2 - y1)**2 + (z2 - z1)**2 \\
        distance = sqrt(distance\_squared) \\
        answer = distance \\
        \\
        Question: How many zeros are at the end of the product 25 \$ times\$ 240? \\
        Code: \\
        def count\_trailing\_zeros\_of\_product(num1, num2): \\
        	\hspace{4ex} product = num1 * num2  \\
        	\hspace{4ex}count\_zeros = 0 \\
        	\hspace{4ex}while product \% 10 == 0: \\
        		\hspace{4ex}\hspace{4ex}count\_\text{zeros} += 1 \\
        		\hspace{4ex}\hspace{4ex}product \//= 10 \\
        	\hspace{4ex}return count\_zeros \\
        num\_zeros = count\_trailing\_zeros\_of\_product(25, 240) \\
        answer = num\_zeros \\
        \\
        If there is even the slightest doubt about solving the problem correctly, output only “Abstain” and do not generate any additional content.\\
        Question: Find the distance between the points \$(2,1,-4)\$ and \$(5,8,-3).\$ \\
        Code: Abstain\\
        \\
        Follow the instructions and handle the problem appropriately without overestimating your abilities.\\
        Question: <question>\\
        Code: \\
    \bottomrule
    \end{tabular}
    \caption{
    Program-of-Thought prompt with an abstention option.}
    \label{tab:prompt_POT}
\end{table*}

\begin{table*}[h]
    \centering
    \small
    \begin{tabular}{p{15cm}}
    \toprule
        Given a question, and the output generated by an LLM in response, please evaluate whether the LLM generated a valid and accurate output for the given question based on the LLM’s output results.\\
        \\
        Indicate the evaluation result after ‘validity’: using one of the following options:\\
        \\
        	•	Answer ‘yes’ if the LLM’s generated output is valid.\\
        	•	Answer ‘no’ if it is invalid.\\
        \\
        The following considerations are important in evaluating the validity of the LLM’s output:\\
        \\
        	1.	Consider whether the LLM’s reasoning, including inference, solution, and code, progresses logically and correctly in the process of deriving the correct answer to the question.\\
        	2.	Assess whether the answer generated by the LLM correctly addresses the given question.\\
        	3.	Even if the output generated by the LLM appears appropriate, evaluate it as "no" if the output and reasoning process are not logically aligned with the question. Alternatively, if an error occurs in the code execution results, always evaluate it as "no".\\
        \\   
        The input format is as follows: \\
        question: The question asked to the LLM. \\
        generated output: The answer generated by the LLM in response to the question.\\
        \\
        Evaluate whether the response generated by the given LLM is valid for the question based on the provided criteria.  \\
        Generate only “yes” or “no,” and do not produce any additional content.\\
        Please make sure to follow below final output format exactly.\\
        validity:\\
    \bottomrule
    \end{tabular}
    \caption{
    Prompt for the iteration baseline and the \ours judge.}
    \label{tab:prompt_judge}
\end{table*}

\section{Qualitative Examples}
We provide examples from MATH and MCQA to illustrate how \ours works in Tables~\ref{tab:caase-MATH} and~\ref{tab:caase-MCQA}.

%\subsubsection{MATH}

\begin{table*}[h]
    \centering
    \small
    \begin{tabular}{p{15cm}}
    \toprule
        \textbf{Question}: A theater group has eight members, of which four are females. How many ways are there to assign the roles of a play that involve one female lead, one male lead, and three different objects that can be played by either gender?\\
        \textbf{Answer}: 1920
        \\
        \textbf{Initial Tier}: GPT-4o\\
        \\
        \textbf{Iteration 1} : GPT-4o\\
        \textcolor{blue}{Generator}:\\
        Code: import math\\
        \hspace{6ex}female\_leads = math.comb(4, 1)\\
        \hspace{6ex}male\_leads = math.comb(4, 1)\\
        \hspace{6ex}objects\_roles = math.comb(6, 3)\\
        \hspace{6ex}total\_ways = female\_leads * male\_leads * objects\_roles\\
        \hspace{6ex}answer = total\_ways\\
        Execute: 320
        \\
        \textcolor{blue}{Judge}: no\\
        \\
        \textbf{Iteration 2} : o1-mini\\
        \textcolor{blue}{Generator}:\\
        Code: import math\\
        \hspace{6ex}female\_leads = 4\\
        \hspace{6ex}male\_leads = 4\\
        \hspace{6ex}objects = math.perm(6, 3)\\
        \hspace{6ex}answer = female\_leads * male\_leads * objects\\
        Execute: 1920\\
        \textcolor{blue}{Judge}: yes\\

    \bottomrule
    \end{tabular}
    \caption{An example of the \ours process in MATH.}
    \label{tab:caase-MATH}
\end{table*}

%\subsubsection{MCQA}
\begin{table*}[h]
    \centering
    \small
    \begin{tabular}{p{15cm}}
    \toprule
        \textbf{Question}: Consider a rhombohedral crystal, with the interatomic distance of 10 Angstrom and the angles $\quad \alpha= \beta= \gamma=30^{\{0\}}$. What is the interplanar distance of the (111) plane of the crystal?\\
        Choices: (A): 8.95 Angstrom, (B) 10.05 Angstrom, (C) 9.54 Angstrom, (D): 9.08 Angstrom\\
        \textbf{Answer}: (C) 9.54 Angstrom
        \\
        \textbf{Initial Tier}: GPT-4o\\
        \\
        \textbf{Iteration 1} : GPT-4o\\
        \textcolor{blue}{Generator}:\\
        First, we use the formula for the interplanar distance in a rhombohedral lattice, given by \textnormal{$d_{hkl} = \frac{a}{\sqrt{h^2 + k^2 + l^2 + 2(hk + hl + kl)\cos\alpha + 3 \cos^2 \alpha - 1}}$}, where \textnormal{$a = 10$} Angstrom, and \textnormal{$\alpha = 30^{\circ}$}. Substituting h, k, l = 1, we get \textnormal{$d_{111} = \frac{10}{\sqrt{1 + 1 + 1 + 2(3)\cos(30^{\circ}) + 3\cos^2(30^{\circ}) - 1}}$} \\
        Calculating the expression inside the square root, we find the denominator to be approximately (10.048), giving ($d_{111}$ approx 9.95 ) Angstrom, which is closest to option (B).\\
        The correct answer is: (B).
        \\
        \textcolor{blue}{Judge}: no\\
        \\
        \textbf{Iteration 2} : o1-mini\\
        \textcolor{blue}{Generator}:\\
        To determine the interplanar distance of the (111) plane in a rhombohedral crystal, we use the formula:\\
        \textnormal{$d_{hkl} = \frac{a}{\sqrt{h^2 + k^2 + l^2 + 2(hk + hl + kl)\cos(\alpha)}}$}\\
        Plugging in the values \textnormal{$a = 10$} \AA, \textnormal{$h = k = l = 1$}, and \textnormal{$\alpha = 30^{\circ}$}:
        \textnormal{$d_{111} = \frac{10}{\sqrt{1 + 1 + 1 + 2(1 + 1 + 1)\cos(30^{\circ})}}$}
        \textnormal{$d_{111} = \frac{10}{\sqrt{3 + 6 \times 0.8660}}$}
        \textnormal{$d_{111} \approx \frac{10}{2.863} \approx 3.49 \text{ \AA}$}.\\
        However, considering crystallographic conventions and the provided answer choices, the closest value is:\\
        The correct answer is: (D)
        \\
        \textcolor{blue}{Judge}: no\\
        \\
        \textbf{Iteration 3} : o1\\
        \textcolor{blue}{Generator}:\\
        The correct answer is: (C)
        \\
        \textcolor{blue}{Judge}: yes\\

    \bottomrule
    \end{tabular}
    \caption{An example of the \ours process in MCQA.}
    \label{tab:caase-MCQA}
\end{table*}

\end{document}